\documentclass{article}

\usepackage[utf8]{inputenc}

\usepackage{etoolbox}
\newcommand{\arxiv}[1]{\iftoggle{neurips}{}{#1}}
\newcommand{\neurips}[1]{\iftoggle{neurips}{#1}{}}
\newtoggle{neurips}
\global\togglefalse{neurips}

\arxiv{
    \usepackage{geometry}
    \geometry{letterpaper, margin=1in}
}

\usepackage{parskip}
\makeatletter
\def\thm@space@setup{%
  \thm@preskip=8pt plus 2pt minus 4pt
  \thm@postskip=\thm@preskip
}
\makeatother

\neurips{
    \usepackage{neurips_2025}

}

\usepackage{txie}

\usepackage{xargs}

\newcommand{\EE}{\mathbb{E}}

\newcommand{\calA}{\mathcal{A}}
\newcommand{\calD}{\mathcal{D}}
\newcommand{\calF}{\mathcal{F}}
\newcommand{\calG}{\mathcal{G}}

\newcommand{\calM}{\mathcal{M}}

\newcommand{\calQ}{\mathcal{Q}}
\newcommand{\calR}{\mathcal{R}}
\newcommand{\calS}{\mathcal{S}}

\newcommand{\unif}{\mathrm{Unif}}
\renewcommand{\epsilon}{\varepsilon}

\newcommand{\pihat}{\widehat{\pi}}

\newcommand{\indep}{\perp\kern-6pt \perp}

\newcommand{\ccov}{C_{\mathrm{cov}}}

\newcommand{\PP}{\mathbb{P}}

\newcommand{\alg}{\mathsf{alg}}

\newcommand{\relu}{\mathrm{ReLU}}

\newcommand{\ta}{\tilde{a}}
\newcommand{\calE}{\mathcal{E}}

\newcommand{\sups}[1]{^{{\scriptscriptstyle#1}}}

\newcommand{\paren}[1]{{\left( #1 \right)}}
\newcommand{\brac}[1]{{\left[ #1 \right]}}

\newcommand{\defeq}{:=}
\newcommandx{\whp}[1][1=\delta]{with probability at least $1-#1$}

\DeclarePairedDelimiter{\set}{\{}{\}}

\newcommand{\TT}{\mathbb{T}}
\newcommand{\R}{\mathbb{R}}

\newcommand{\eps}{\varepsilon} 

\newcommand{\RNum}[1]{\uppercase\expandafter{\romannumeral #1\relax}}

\newcommand{\cA}{\mathcal{A}}

\newcommand{\cD}{\mathcal{D}}
\newcommand{\cE}{\mathcal{E}}
\newcommand{\cF}{\mathcal{F}}
\newcommand{\cG}{\mathcal{G}}
\newcommand{\cH}{\mathcal{H}}
\newcommand{\cI}{\mathcal{I}}

\newcommand{\cL}{\mathcal{L}}
\newcommand{\cM}{\mathcal{M}}

\newcommand{\cQ}{\mathcal{Q}}
\newcommand{\cR}{\mathcal{R}}
\newcommand{\cS}{\mathcal{S}}
\newcommand{\cT}{\mathcal{T}}

\newcommand{\cX}{\mathcal{X}}
\newcommand{\cY}{\mathcal{Y}}

\newcommand{\lr}{\langle}
\newcommand{\rr}{\rangle}
\newcommand{\ind}[1]{\sups{(#1)}}
\newcommand{\abs}[1]{\left|#1\right|}

\newcommand{\cond}[2]{\left[\left.#1\right|#2\right]}
\newcommand{\nrm}[1]{\left\|#1\right\|}
\newcommand{\linf}[1]{\nrm{#1}_{\infty}}
\newcommand{\filt}{\mathfrak{F}}
\newenvironment{proofof}[1]{
\paragraph{Proof of #1} 
}{
\qed
}
\newcommand{\epsapp}{\epsilon_{\rm app}}

\newcommand{\id}{\mathbf{I}}

\newcommand{\Bern}[1]{\mathrm{Bern}\paren{#1}}
\newcommand{\Cmp}{\mathsf{C}}

\newcommand{\Rs}{R^\star}
\newcommand{\Rsp}{R^{\sharp}}
\newcommand{\Qs}{Q^\star}
\newcommand{\Qsp}{Q^{\sharp}}
\newcommand{\Vs}{V^\star}
\newcommand{\Vsp}{V^{\sharp}}
\newcommand{\pis}{\pi^\star}

\newcommand{\Mstar}{M^\star}

\newcommand{\piexp}{\pi_{\rm ref}}

\newcommandx{\LBEh}[2][1={\cD},2=h]{\cE_{#1,#2}}
\newcommandx{\LBEt}[2][1=t,2=h]{\cE_{\cD^\iter{#1},#2}}
\newcommandx{\LBE}[1][1={\cD}]{\cL^{\mathsf{BE}}_{#1}}
\newcommandx{\LDBE}[1][1={\cD}]{\cL^{\mathsf{BR}}_{#1}}
\newcommandx{\LRM}[1][1={\cD}]{\cL^{\sf RM}_{#1}}
\newcommandx{\LPBRM}[1][1={\cD}]{\cL^{\sf PbRM}_{#1}}
\newcommandx{\BE}{\cT^\star}
\newcommandx{\BER}[1][1=R]{\cT_{#1}}
\newcommandx{\BERh}[2][1=R,2=h]{\cT_{#1,#2}}
\newcommand{\bigO}[1]{O\paren{#1}}
\newcommand{\tbO}[1]{\widetilde{O}\paren{#1}}
\newcommand{\omu}{\overline{\mu}}

\usepackage{makecell}

\usepackage{prettyref}
\newcommand{\pref}[1]{\prettyref{#1}}

\newcommand{\savehyperref}[2]{\texorpdfstring{\hyperref[#1]{#2}}{#2}}
\newrefformat{eq}{\savehyperref{#1}{Eq.~\textup{(\ref*{#1})}}}
\newrefformat{eqn}{\savehyperref{#1}{Equation~\ref*{#1}}}
\newrefformat{lem}{\savehyperref{#1}{Lemma~\ref*{#1}}}
\newrefformat{lemma}{\savehyperref{#1}{Lemma~\ref*{#1}}}
\newrefformat{def}{\savehyperref{#1}{Definition~\ref*{#1}}}
\newrefformat{line}{\savehyperref{#1}{Line~\ref*{#1}}}
\newrefformat{thm}{\savehyperref{#1}{Theorem~\ref*{#1}}}
\newrefformat{corr}{\savehyperref{#1}{Corollary~\ref*{#1}}}
\newrefformat{cor}{\savehyperref{#1}{Corollary~\ref*{#1}}}
\newrefformat{sec}{\savehyperref{#1}{Section~\ref*{#1}}}
\newrefformat{subsec}{\savehyperref{#1}{Section~\ref*{#1}}}
\newrefformat{app}{\savehyperref{#1}{Appendix~\ref*{#1}}}
\newrefformat{ass}{\savehyperref{#1}{Assumption~\ref*{#1}}}
\newrefformat{asmp}{\savehyperref{#1}{Assumption~\ref*{#1}}}
\newrefformat{ex}{\savehyperref{#1}{Example~\ref*{#1}}}
\newrefformat{fig}{\savehyperref{#1}{Figure~\ref*{#1}}}
\newrefformat{alg}{\savehyperref{#1}{Algorithm~\ref*{#1}}}
\newrefformat{rem}{\savehyperref{#1}{Remark~\ref*{#1}}}
\newrefformat{conj}{\savehyperref{#1}{Conjecture~\ref*{#1}}}
\newrefformat{prop}{\savehyperref{#1}{Proposition~\ref*{#1}}}
\newrefformat{proto}{\savehyperref{#1}{Protocol~\ref*{#1}}}
\newrefformat{prob}{\savehyperref{#1}{Problem~\ref*{#1}}}
\newrefformat{claim}{\savehyperref{#1}{Claim~\ref*{#1}}}
\newrefformat{que}{\savehyperref{#1}{Question~\ref*{#1}}}
\newrefformat{op}{\savehyperref{#1}{Open Problem~\ref*{#1}}}
\newrefformat{fn}{\savehyperref{#1}{Footnote~\ref*{#1}}}
\newrefformat{tab}{\savehyperref{#1}{Table~\ref*{#1}}}
\newrefformat{fig}{\savehyperref{#1}{Figure~\ref*{#1}}}
\newrefformat{proc}{\savehyperref{#1}{Procedure~\ref*{#1}}}
\newrefformat{property}{\savehyperref{#1}{Property~\ref*{#1}}}

\let\oldparagraph\paragraph
\renewcommand{\paragraph}[1]{\oldparagraph{#1.}}

\usepackage[suppress]{color-edits}

\addauthor[Tengyang]{tx}{teal}
\addauthor[Zeyu]{zj}{orange}
\addauthor[Fan]{cf}{blue}
\addauthor{sr}{magenta}

\usepackage[T1]{fontenc}
\usepackage[scaled=.91]{helvet}
\usepackage{inconsolata}
\usepackage{microtype}

\allowdisplaybreaks

\title{Outcome-Based Online Reinforcement Learning: Algorithms and Fundamental Limits}

\neurips{
\author{}
}

\arxiv{
\author{Fan Chen\footnote{Massachusetts Institute of Technology.}\\{\small \texttt{fanchen@mit.edu}} \and
    Zeyu Jia\footnotemark[1]\\{\small \texttt{zyjia@mit.edu}} \and Alexander Rakhlin\footnotemark[1]\\{\small \texttt{rakhlin@mit.edu}} \and  Tengyang Xie\footnote{University of Wisconsin–Madison.}\\{\small \texttt{tx@cs.wisc.edu} }
}
}

\date{\today}

\begin{document}

\maketitle

\begin{abstract}
Reinforcement learning with outcome-based feedback faces a fundamental challenge: when rewards are only observed at trajectory endpoints, how do we assign credit to the right actions? This paper provides the first comprehensive analysis of this problem in online RL with general function approximation. We develop a provably sample-efficient algorithm achieving $\widetilde{O}({C_{\rm cov} H^3}/{\epsilon^2})$ sample complexity, where $C_{\rm cov}$ is the coverability coefficient of the underlying MDP. By leveraging general function approximation, our approach works effectively in large or infinite state spaces where tabular methods fail, requiring only that value functions and reward functions can be represented by appropriate function classes. Our results also characterize when outcome-based feedback is statistically separated from per-step rewards, revealing an unavoidable exponential separation for certain MDPs. For deterministic MDPs, we show how to eliminate the completeness assumption, dramatically simplifying the algorithm. We further extend our approach to preference-based feedback settings, proving that equivalent statistical efficiency can be achieved even under more limited information. Together, these results constitute a theoretical foundation for understanding the statistical properties of outcome-based reinforcement learning.

\end{abstract}

\section{Introduction}

Reinforcement learning with outcome-based feedback is a fundamental paradigm where agents receive rewards only at the end of complete trajectories rather than at individual steps. This feedback model naturally arises in many applications, from large language model training \citep{ouyang2022training,bai2022training,jaech2024openai}, where human preferences are provided for entire outputs rather than individual tokens, to clinical trials, where patient outcomes are only observable after a complete treatment regimen. Despite the prevalence of such settings, the statistical implications of outcome-based feedback for online exploration remain poorly understood.

In traditional reinforcement learning \citep{sutton1998reinforcement}, agents observe rewards immediately after each action, providing a granular signal that directly links actions to their consequences. In contrast, outcome-based feedback presents a fundamental challenge: when rewards are only observed at the trajectory level, determining which specific actions contributed to the final outcome becomes significantly more difficult. This credit assignment problem is particularly acute in sequential decision-making tasks with long horizons, where many different action combinations could lead to the observed outcome.

While recent work \citep{jia2025we} has shown that outcome-based feedback is sufficient for offline reinforcement learning under certain conditions, the feasibility of efficient online exploration with only trajectory-level feedback remains an open question. Online learning—where an agent actively explores to gather new data—is essential for adaptive systems that must learn in dynamic environments without pre-collected datasets. This leads to our central question:

\begin{quote}
\centering
\emph{When is online exploration with outcome-based reward statistically tractable?}
\end{quote}

This question has been studied in the setting where the reward function is assumed to be well-structured~\citep{efroni2021reinforcement,pacchiano2021dueling,chatterji2021theory,cassel2024near,lancewicki2025near}, with a primary focus on the \emph{linear} reward functions. Similar reliance on the well-behaved reward structure also appears in the recent work on Reinforcement Learning from Human Feedback (RLHF)~\citep{chen2022human,chen2022unified,wu2023making,wang2023rlhf}, where only \emph{preference} feedback is available.
However, well-behaved reward structure is dedicated and might fail to capture many real-world scenarios with general function approximation. \cfcomment{added}
In this paper, we address this question by providing a comprehensive theoretical analysis of outcome-based online reinforcement learning with general function approximation. We investigate when efficient exploration is possible with only trajectory-level feedback and characterize the fundamental statistical limits of learning in this setting. Our main results are as follows:

\begin{enumerate}[(1),leftmargin=*]
\item We present a model-free algorithm for outcome-based online RL with general function approximation (\cref{alg: model-free}) that relies solely on trajectory-level reward feedback rather than per-step feedback. Our algorithm achieves a complexity bound of $\widetilde{O}({\ccov H^3}/{\epsilon^2})$ under standard realizability and completeness assumptions, where $\ccov$ is the coverability coefficient that measures an intrinsic complexity of the underlying MDP. %
This bound applies in the general function approximation setting where state spaces may be large or infinite, requiring only that value functions can be represented by an appropriate function class with bounded statistical complexity. 

\item For the special case of deterministic MDPs, we present a simpler algorithm based on Bellman residual minimization (\cref{alg: deterministic}) that achieves similar theoretical guarantees with improved computational efficiency. %

\item As extension, we generalize our approach to preference-based reinforcement learning (\cref{sec: PbRL}), where feedback comes in the form of binary preferences between trajectory pairs under the Bradley-Terry-Luce model. 
This extension bridges the gap to practical reinforcement learning from human feedback (RLHF) scenarios, where even outcome reward feedback is rare. %

\item We also identify a fundamental separation between outcome-based and per-step feedback (\cref{sec: lower-bound}). Specifically, there exists a MDP with known transition dynamics and horizon $H=2$, and the reward being a $d$-dimensional generalized linear function, while in this problem $e^{\Omega(d)}$ samples are necessary to learn a near-optimal policy with only outcome reward. However, such a problem is known to be \emph{easy} with per-step reward feedback, in the sense that existing algorithms can return an $\epsilon$-optimal within $\tbO{d^2/\epsilon^2}$ rounds with per-step feedback. 
This separation demonstrates that delicate analysis based on well-behaved reward structure can fail catastrophically when only outcome reward feedback is available. %

\end{enumerate}

Our results provide a theoretical foundation for understanding when outcome-based exploration is tractable and when it presents insurmountable statistical barriers. By characterizing these fundamental limits, we offer guidance for the development of efficient algorithms for learning from trajectory-level feedback in online settings and highlight the precise conditions under which outcome-based feedback is statistically equivalent to per-step feedback.

\section{Preliminaries}

\paragraph{Markov Decision Process}
An MDP $M$ is specified by a tuple $(\cS,\cA,\TT,\rho,R,H)$, with state space $\cS$, action space $\cA$, horizon $H$, transition kernel $\TT=(\TT_h:\cS\times\cA\to\Delta(\cS))_{h=1}^{H-1}$, is initial state distribution $\rho\in\Delta(\cS)$, and the mean reward function $R=(R_h:\cS\times\cA\to [0,1])_{h=1}^H$. 
At the start of each episode, the environment randomly draws an initial state $s_1\sim \rho$, and then at each step $h\in[H]$, after the agent takes action $a_h$, the environment generates the next state $s_{h+1}\sim \TT(\cdot|s_h,a_h)$. The episode terminates immediately after $a_{H}$ is taken, and, for notational simplicity, we denote $s_{H+1}$ to be the deterministic terminal state. We denote $\tau=(s_1,a_1,\cdots,s_H,a_H)$ to be the trajectory, and throughout this paper we always assume the reward function is normalized, i.e., $R(\tau)\defeq \sum_{h=1}^H R_h(s_h,a_h)\in[0,1]$ almost surely.

In addition to the states, the learner may also observe the reward feedback after the episode terminates. In the \emph{process reward} feedback setting, the learner receives a random reward vector $(r_1,\cdots,r_H)\in[0,1]^H$ such that $\EE[r_h|\tau]=R_h(s_h,a_h)$ for each $h\in[H]$. In the \emph{outcome reward} setting, the learner only receives a single reward value $r\in[H]$ such that $\EE[r|\tau]=\sum_{h=1}^H R_h(s_h,a_h)$.

\paragraph{Policies, value functions, and the Bellman operator}
A (randomized) policy $\pi$ is specified as $\set{\pi_h:\cS\to\Delta(\cA)}$, and it induces a distribution $\PP^\pi$ of trajectory $\tau=(s_1,a_1,\cdots,s_H,a_H)$ by $s_1\sim \rho$, and for each $h\in[H]$, $a_h\sim \pi_h(s_h)$, $s_{h+1}\sim \TT_h(s_h,a_h)$. We let $\EE^\pi[\cdot]$ to be the corresponding expectation. 

The expected cumulative reward of a policy $\pi$ is given by $J(\pi)\defeq \EE^{\pi}\brac{\sum_{h=1}^H R_h(s_h,a_h)}$. The value function and $Q$-function of $\pi$ is defined as
\begin{align*}
    V_h^\pi(s)\defeq \EE^{\pi}\cond{\sum_{\ell=h}^H R_\ell(s_\ell,a_\ell)}{s_h=s}, \qquad 
    Q^\pi_h(s,a)\defeq \EE^{\pi}\cond{\sum_{\ell=h}^H R_\ell(s_\ell,a_\ell)}{s_h=s,a_h=a}.
\end{align*}
Let $\pis$ denote an optimal policy (i.e., $\pis\in\argmax_\pi J(\pi)$), and let $\Vs$ and $\Qs$ be the corresponding value function and $Q$-function. It is well-known that $(\Vs,\Qs)$ satisfies the following Bellman equation for each $s\in\cS, a\in\cA, h\in[H]$:
\begin{align}\label{eq:Bellman-eq}
    \Vs_h(s)=\max_{a\in\cA} \Qs_h(s,a), \qquad
    \Qs_h(s,a)=R_h(s,a)+\EE_{s'\sim \TT_h(\cdot|s,a)} \Vs_{h+1}(s'),
\end{align}
with the convention that $\Vs_{H+1}=0$.
Therefore, we define the Bellman operator $\cT_h$ as follows: for any $f:\cS\times\cA\to\R$, $\cT_h f$ is defined as
\begin{align}\label{eq:Bellman-op}
    [\cT_hf](s,a)\defeq R_h(s,a)+\EE_{s'\sim \TT_h(\cdot|s,a)} \max_{a'\in\cA} f(s',a').
\end{align}
Then, it is straightforward to verify that the Bellman equation reduces to $\Qs_h=\cT_h\Qs_{h+1}$ for $h\in[H]$.

\paragraph{Complexity measure of the MDP}
Coverability is a natural notion for measuring the difficulty of learning in the underlying MDP~\citep{xie2022role}. 
\begin{definition}[Coverability]\label{def: coverability}
    For a given MDP $M$ and a policy class $\Pi$, the coverability $\ccov$ is defined as %
    $$\ccov (\Pi;M)\coloneqq \min_{\mu_1,\cdots,\mu_H\in\Delta(\cS\times\cA)}\max_{h\in [H],\pi\in\Pi}\linf{\frac{d^\pi_h}{\mu_h}},$$
    where $\linf{\frac{d^\pi_h}{\mu_h}}\defeq \max_{s\in\cA,a\in\cA} \frac{d^\pi_h(s,a)}{\mu_h(s,a)}$. \cfcomment{changed def'n}
\end{definition}
The coverability coefficient of an MDP is an inherent measure of the diversity of the state-action distributions. Our main upper bounds scale with the coverability of the underlying MDP $\Mstar$, and in this case we abbreviate $\ccov(\Pi)\defeq \ccov(\Pi;\Mstar)$ for succinctness.

\newcommand{\linfn}[1]{\|#1\|_{\infty}}

\paragraph{Function approximation}
In this paper, we work with (model-free) function approximation, where the learner have access to a \emph{value function class} $\cF=\cF_1\times\cdots\times\cF_H$ and a \emph{reward function class} $\cR=\cR_1\times\cdots\times\cR_H$ with each $\cF_h, \cR_h\subseteq (\cS\times\cA\to [0,1])$. 

The function class $\cF$ and $\cR$ consist of candidate functions to approximate $\Qs$ and the ground-truth reward function $\Rs$.\footnote{In the following, we always write $\Rs$ for the true reward function to avoid confusion.} In the literature of RL with general function approximation, it is typically assumed that the function classes are \emph{realizable}, i.e., $\Qs\in\cF$ and $\Rs\in\cR$. In this paper, we adopt the following relaxed realizability condition with a fixed approximation error $\epsapp\geq 0$.
\begin{assumption}[Realizability]\label{asmp:realizable}
There exists $\Qsp\in\cF$ and $\Rsp\in\cR$ such that $\max_{h\in[H]}\linfn{\Qsp_h-\Qs_h}\leq \epsapp$, $\max_{h\in[H]}\linfn{\Rsp_h-\Rs_h}\leq \epsapp$.
\end{assumption}

For each value function $f\in\cF$, it induces a greedy policy $\pi_f$ given by $\pi_{f,h}(s)\defeq \argmax_{a\in\cA} f(s,a)$. Therefore, the value function class $\cF$ induces a policy class $\Pi_\cF\defeq \set{\pi_f: f\in\cF}$, and we take our policy class $\Pi=\Pi_\cF$ for the remaining part of this paper. 

The complexity of the function class is measured by the covering number.
\begin{definition}[Covering number]\label{def:covering}
For a function class $\cH\subseteq (\cX\to\R)$ and parameter $\alpha\geq 0$, an $\alpha$-covering of $\cH$ (with respect to the sup norm) is a subset $\cH'\subseteq \cH$ such that for any $f\in\cH$, there exists $f'\in\cH'$ with $\sup_{x\in\cX}|f(x)-f'(x)|\leq \alpha$. We define the $\alpha$-covering number of $\cH$ as $N(\cH,\alpha)\defeq \min\set{|\cH'|: \cH'\text{ is a $\alpha$-covering of }\cH}$.
\end{definition}

\paragraph{Bellman completeness}
A reward function $R=(R_1,\cdots,R_H)\in\cR$ induces a Bellman operator as
\begin{align*}
    [\cT_{R,h}f](s,a)\defeq R_h(s,a)+\EE_{s'\sim \TT_h(\cdot|s,a)} \max_{a'\in\cA} f_{h+1}(s',a'), \qquad \forall f=(f_1,\cdots,f_H)\in\cF,
\end{align*}
where we also adopt the notation $f_{H+1}=0$ for any $f\in\cF$. Most literature on RL with general function approximation also makes use of a richer comparator function class $\cG=\cG_1\times\cdots\times\cG_H$ that satisfies the following \emph{Bellman completeness}~\citep[etc.]{jin2021bellman,xie2021bellman,xie2022role}. 
\begin{assumption}[Bellman completeness]\label{asmp:complete}
For each $h\in[H]$, $\cF_h\subseteq \cG_h$. For any $f\in\cF$ and $R\in\cR$, it holds $\inf_{g_h\in\cG_h}\linf{\cT_{R,h} f_{h+1}-g_h}\leq \epsapp$ for $h\in[H]$.
\end{assumption}

\paragraph{Miscellaneous notation}
For any $p\in[0,1]$, we define $\Bern{p}$ to be the Bernoulli distribution with $\PP(X=1)=p$. For functions $f$ and $g\geq 0$, we use $f = O(g)$ to denote that there exists a universal constant $C$ such that $f\le C\cdot g$.

\section{Sample-Efficient Online RL with Outcome Reward}\label{sec: upper-bound}

In this section, we present a model-free RL algorithm with outcome reward, which achieves sample complexity guarantee scaling with the coverability coefficient and the log-covering number of the function classes.

\subsection{Main Result}\label{sec: upper-bound-cov}

We present \cref{alg: model-free}, which is based on the principle of optimism. For simplicity of presentation, we assume that the initial state $s_1$ is fixed.

The crux of the proposed algorithm is a new method for performing Fitted-Q Iteration with only outcome reward, in contrast to most existing RL algorithms (with general function approximation) that make use the process reward $(r_1,\cdots,r_H)$ to fit the Q-function for each step~\citep[etc.]{du2021bilinear,jin2021bellman}. A natural first idea is to fit, given a dataset $\cD=\set{(\tau,r)}$ consisting of previously observed (trajectory, outcome reward) pairs, a reward model from the reward function class $\cR$ by optimizing the following reward model loss:
\begin{align}\label{eq:loss-RM}
    \LRM(R)\defeq \sum_{(\tau,r)\in\cD} \paren{ \sum_{h=1}^H R_h(s_h,a_h) - r }^2.
\end{align}
As discussed below in \cref{rem:separate_opt}, directly fitting an estimated reward model based on $\LRM$ can lead to bad performance. Instead, our algorithm jointly optimizes over the value functions and reward models, as detailed below.

For any proxy reward model $R\in\cR$ and a value function $f\in\cF$, we define the Bellman error at step $h\in[H]$ as
\begin{align}\label{eq:loss-Bellman-error}
    \LBEh(f_h,f_{h+1};R)\defeq \sum_{(\tau,r)\in\cD} \left( f_h(s_h,a_h) - R_h(s_h,a_h) - \max_{a'} f_{h+1}(s_{h+1},a') \right)^2,
\end{align}
a measure of violation of the Bellman equation \eqref{eq:Bellman-eq} with the proxy reward model $R$. 
Then, we introduce the Bellman loss defined as
\begin{align}\label{eq:loss-Bellman-full}
    \LBE(f;R)\defeq \sum_{h=1}^H \LBEh(f_h,f_{h+1};R) - \inf_{g\in\cG} \sum_{h=1}^H \LBEh(g_h, f_{h+1}; R), 
\end{align}
where we subtract the infimum of $g\in\cG$ over the helper function class $\cG$, a common approach to overcoming the double-sampling problem~\citep{antos2008learning,zanette2020learning,jin2021bellman}.

\begin{algorithm}[t]
\caption{Outcome-Based Exploration with Optimism}\label{alg: model-free}
\label{alg:golf}
{\bfseries input:} Q-function class $\Fcal$, reward function class $\cR$, comparator class $\cG$, \ parameter $\lambda>0$, reference policy $\piexp$. \\
{\bfseries initialize:} $\Dcal \leftarrow \emptyset$. %
\begin{algorithmic}[1]
\For{$t = 1,2,\dotsc,T$}
\State Compute the optimistic estimates:
\begin{align*}
(f^\iter{t},R^\iter{t})=\max_{f \in \cF, R \in \cR} \lambda f_1(s_1) - \LBE(f;R) - \LRM(R).
\end{align*}
    \State Select policy $\pi^\iter{t} \leftarrow \pi_{f^\iter{t}}$. 
    \For{$h=1,2,\cdots,H$}
    \State Execute $\pi^\iter{t}\circ_h \piexp$ for one episode and obtain $(\tau^\iter{t,h},r^\iter{t,h})$
    \State Update dataset: $\Dcal \leftarrow \Dcal \cup \set{(\tau^\iter{t,h},r^\iter{t,h})}$.
    \EndFor
    
\EndFor
\State Output $\pihat = \unif(\pi^\iter{1:T})$.
\end{algorithmic}
\end{algorithm}

\paragraph{Algorithm} First fix an arbitrary policy $\piref$ (can be the policy which takes an arbitrary action $a_0$ at al states). The proposed algorithm takes in a value function class $\cF$, a reward function class $\cR$ and a comparator function class $\cG$, and performs the following two steps for each iteration $t=1,2,\cdots,T$: 
\begin{enumerate}[leftmargin=15pt]
\item (Optimism) Compute optimistic estimates of $(\Qs,\Rs)$ through solving the following joint maximization problem with the dataset $\cD$ consisting of all previously observed (trajectory, outcome reward) pairs:
\begin{align}\label{eq:max-f-R}
    (f^\iter{t},R^\iter{t})=\max_{f \in \cF, R \in \cR} \lambda f_1(s_1) - \LBE(f;R) - \LRM(R),
\end{align}
where for any $f\in\cF$ we denote $f_1(s_1)\defeq \max_{a\in\cA} f_1(s_1,a)$ to be the value of $f$ at the initial state. Therefore, the optimization problem~\eqref{eq:max-f-R} enforces optimism by balancing the estimated value $f_1(s_1)$ and the estimation error $\LBE(f;R) + \LRM(R)$ though a hyper-parameter $\lambda\geq 0$.

\item (Data collection) Based on the optimism estimate $f^\iter{t}$, the algorithm selects $\pi^\iter{t}\defeq \pi_{f^\iter{t}}$. To collect data, the algorithm then executes the exploration policies $\pi\circ_h \piexp$ for each $h\in[H]$, where for any policy $\pi$ and $\piexp$, we let $\pi\circ_h \piexp$ be the policy that executes $\pi$ for the first $h$ steps, and then executes $\piexp$ starting at the $(h+1)$-th step.
\end{enumerate}

\newcommand{\NT}[1]{N_{#1}}
\newcommand{\sset}[1]{\left\{#1\right\}}
\paragraph{Theoretical analysis}
For \pref{alg: model-free}, we provide the following sample complexity guarantee, which scales with the coverability $\ccov=\ccov(\Pi_\cF)$, where $\Pi_\cF=\set{\pi_f:f\in\cF}$ is the policy class induced by $\cF$. To simplify the presentation, we denote $\log \NT{T}\defeq \inf_{\alpha\geq 0}\paren{ \log N(\alpha)+T\alpha }$, where $N(\alpha)$ is defined as
\begin{align*}
    N(\alpha)\defeq \max_{h\in[H]}\set{ N(\cF_h,\alpha), N(\cR_h,\alpha), N(\cG_h,\alpha) }.
\end{align*}
With the function classes being parametric, it is clear that $\log\NT{T}\leq \bigO{d\log(T)}$. %
\begin{theorem}\label{thm: model-free-thm}
Let $\delta\in(0,1)$. Suppose that \cref{asmp:realizable} and \pref{asmp:complete} hold, and the parameters are chosen as
\begin{align}\label{eq: model-free-thm-specs}
    \lambda=c_0\max\sset{ \frac{H\log(\NT{TH^2}/\delta)}{\eps}, TH\epsapp }, \qquad 
    T\geq c_1\frac{\ccov H^2 \log(T)}{\eps^2}\cdot \log(\NT{TH^2}/\delta),
\end{align}
where $c_0, c_1>0$ are absolute constants.
Then \whp, the output policy $\pihat$ of \cref{alg: model-free} satisfies $\Vs(s_1)-V^{\pihat}(s_1)\leq \eps+\bigO{\ccov H^2\log(T)\cdot \epsapp}$. 
\end{theorem}
The proof of \pref{thm: model-free-thm} is deferred to \pref{app: upper-bound-cov}. Particularly, we note that when the function classes satisfy $\log\NT{T}\leq \tbO{d}$ and $\epsapp=0$, \cref{alg: model-free} outputs an $\epsilon$-optimal policy with sample complexity
\begin{align*}
    TH\leq\tbO{\frac{\ccov dH^3 }{\eps^2}}.
\end{align*}
Notably, the coverability $\ccov$ measures the inherent complexity of the underlying MDP $\Mstar$~\citep{xie2022role} and it is independent of the reward function class. As our result only depends on the coverability $\ccov$ and the statistical complexity of the function classes, it does \emph{not} rely on the structure of reward functions, while previous works assume the reward functions are either linear~\citep{efroni2021reinforcement,cassel2024near}
or admit low \emph{trajectory} eluder dimension~\citep{chen2022human,chen2022unified}.%

\begin{remark}
\label{rem:separate_opt}
    In \pref{alg: model-free}, the reward functions $R\ind{t}$ and $Q$-functions $f\ind{t}$ are jointly optimized (see \pref{eq:max-f-R}). A natural question is whether these can be optimized separately—i.e., first learning a fitted reward model and then applying optimism to the $Q$-functions based on the learned reward model. We show that this decoupled approach can lead to failures: due to reward model mismatch, the algorithm may become `trapped' in regions where the exploratory policy fails to gather informative data. As a result, the sample complexity can become infinite in the worst case. See \pref{sec: hard-case-separate} in the appendix for details.
\end{remark}

\subsection{A Simpler Algorithm for Deterministic MDPs}

A disadvantage of \cref{alg: model-free} is that it requires solving a max-min optimization problem~\eqref{eq:max-f-R}, as the Bellman loss $\LBE$ involves a minimization problem over $\cG$. While such computationally inefficient optimization problems are the common subroutines of existing function approximation RL algorithms~\citep[etc.]{jin2021bellman,foster2021statistical,foster2022complexity,chen2022unified}, it turns out that \cref{alg: model-free} can be significantly simplified when the transition dynamics in underlying MDP are deterministic.

\begin{assumption}\label{asmp:deterministic}
The transition kernel $\TT$ is deterministic, i.e., for any $h\in [H]$ and $s_h\in \calS, a_h\in \calA$, there is a unique state $s_{h+1}\in \calS$ such that $\TT_h(s_{h+1}\mid s_h, a_h)=1$.
\end{assumption}
Note that in this setting, the initial state $s_1$ and the outcome reward $r$ can still be random. This setting is also referred to as \emph{Deterministic Contextual MDP} in \citet{xie2024exploratory}.

\paragraph{Value difference as reward model}
A key observation is that, when the underlying MDP $\Mstar$ is deterministic, the Bellman equation~\eqref{eq:Bellman-eq} trivially reduces to the following equality
\begin{align*}
    \Qs_h(s_h,a_h)=\Rs_h(s_h,a_h)+\Vs_{h+1}(s_{h+1}),
\end{align*}
which holds almost surely. Hence, for any trajectory $\tau$, it holds that
\begin{align*}
    \Rs(\tau)=\sum_{h=1}^H \Rs_h(s_h,a_h)=\sum_{h=1}^H \brac{ \Qs_h(s_h,a_h)-\Vs_{h+1}(s_{h+1}) }.
\end{align*}
Therefore, any value function $f\in\cF$ induces an outcome reward model $R^f:(\cS\times\cA)^H\to \R$ defined as 
\begin{align*}
    R^f(\tau)=\sum_{h=1}^H \brac{ f_h(s_h,a_h)-f_{h+1}(s_{h+1}) },
\end{align*}
where we adopt the notation  $f_h(s)\defeq \max_{a\in\cA} f_h(s,a)$ for $h\in[H]$. This observation motivates the following Bellman Residual loss:
\begin{align}\label{eq:loss-BR}
    \LDBE(f)\defeq \sum_{(\tau,r)\in\cD}  \paren{ \sum_{h=1}^H \brac{ f_h(s_h,a_h) - f_{h+1}(s_{h+1}) } - r }^2, 
\end{align}
where $\cD=\set{(\tau,r)}$ is any dataset consisting of (trajectory, outcome reward) pairs.

\begin{algorithm}[t]
    \caption{Outcome-Based Exploration with Optimism for Determinsitic MDP}
    \label{alg: deterministic}
    {\bfseries input:} Function class $\Fcal$, parameter $\lambda>0$. \\
    {\bfseries initialize:} $\Dcal \leftarrow \emptyset$, initial estimate $f^\iter{1}\in\cF$. 
    \begin{algorithmic}[1]
    \For{$t = 1,2,\dotsc,T$}
        \State Receive $s^\iter{t}$ and compute the optimistic estimates:
            \begin{align*}
            f^\iter{t}=\max_{f \in \cF} \lambda f_1(s_1^\iter{t}) - \LDBE(f).
            \end{align*}
        \State Select policy $\pi^\iter{t} \leftarrow \pi_{f^\iter{t}}$. 
        \State Execute $\pi^\iter{t}$ to obtain a trajectory $\tau^\iter{t}=(s_1^\iter{t},a_1^\iter{t},\ldots,s_H^\iter{t},a_H^\iter{t})$ with outcome reward $r^\iter{t}$.
        \State Update dataset: $\Dcal \leftarrow \Dcal \cup \set{(\tau^\iter{t},r^\iter{t})}$.
    \EndFor
    \State Output $\pihat = \unif(\pi^\iter{1:T})$.
    \end{algorithmic}
\end{algorithm}

\paragraph{Bellman Residual Minimization (BRM) with Optimism} For deterministic MDP, we propose \cref{alg: deterministic} as a simplification of our main algorithm. Similar to \cref{alg: model-free}, the proposed algorithm takes in the value function class $\cF$ and alternates between the following two steps for each round $t=1,2,\cdots,T$: 
\begin{enumerate}[leftmargin=15pt]
\item (Optimism) Compute optimistic estimates of $\Qs$ through solving the following maximization problem with the dataset $\cD$ consisting of all previously observed (trajectory, outcome reward) pairs:
\begin{align}\label{eq:max-f-DMDP}
    f^\iter{t}=\max_{f \in \cF} \lambda f_1(s_1) - \LDBE(f),
\end{align}
enforcing optimism by balancing the estimated value $f_1(s_1)$ and the Bellman residual loss $\LDBE(f)$.

\item (Data collection) Based on the optimistic estimate $f^\iter{t}$, selects $\pi^\iter{t}\defeq \pi_{f^\iter{t}}$ and collect a trajectory $\tau^\iter{t}=(s_1^\iter{t},a_1^\iter{t},\ldots,s_H^\iter{t},a_H^\iter{t})$ with outcome reward $r^\iter{t}$. 
\end{enumerate}

Compared to \pref{alg: model-free}, \pref{alg: deterministic} has the several advantages. First, it does not rely on the reward function class $\cR$ and the comparator function class $\cG$, and the Bellman residual loss $\LDBE$ is much simpler than the Bellman loss $\LBE$, thanks to the deterministic nature of the underlying MDP. Therefore, \cref{alg: deterministic} is more amenable to computationally efficient implementation, as it replaces the max-min optimization problem~\eqref{eq:max-f-R} in \cref{alg: model-free} with a much simpler maximization problem~\eqref{eq:max-f-DMDP}.
Further, for every round $t$, the algorithm only needs to collect one episode from the greedy policy $\pi^\iter{t}$. %

\paragraph{Theoretical analysis}
We present the upper bound of \cref{alg: deterministic} in terms of the following notion of coverability,
\begin{align*}
    \ccov'(\Pi)\defeq \EE_{s_1\sim \rho} \ccov(\Pi;\Mstar_{s_1}),
\end{align*}
where $\Mstar$ is the underlying MDP, $\Mstar_{s_1}$ is the MDP with deterministic initial state $s_1$ and the same transition dynamics as $\Mstar$, and $\Pi=\Pi_{\cF}$ is the policy class induced by $\cF$. In general, $\ccov'(\Pi)$ is always an upper bound on the coverability $\ccov(\Pi)$, and the guarantee of \cref{alg: deterministic} scales with $\ccov'(\Pi)$ as it avoids the layer-wise exploration strategy of \cref{alg: model-free}. We also denote
\begin{align*}
    \log\NT{\cF,T}\defeq \inf_{\alpha\geq 0}\paren{ \max_{h\in[H]}N(\cF_h,\alpha)+T\alpha }.
\end{align*}

\begin{theorem}\label{thm: deterministic}
Let $\delta\in(0,1)$. Suppose that \cref{asmp:realizable} holds, and the parameters are chosen as
\begin{align}\label{eq: deterministic-specs}
    \lambda=c_0\max\sset{ \frac{H^3\log(\NT{\cF,T}/\delta)}{\eps}, T\epsapp }, \quad
    T\geq c_1\frac{\ccov'(\Pi) H^4 \log(T)}{\eps^2}\cdot \log(\NT{\cF,T}/\delta),
\end{align}
where $c_0, c_1>0$ are absolute constants. Then \whp, \pref{alg: deterministic} achieves
\begin{align*}
    \frac1T \sum_{t=1}^T\paren{ \Vs(s_1^\iter{t})-V^{\pi^\iter{t}}(s_1^\iter{t}) } 
    \leq&~ \eps+\bigO{\ccov'(\Pi)H\log(T)\cdot \epsapp}.
\end{align*}
\end{theorem}

The above upper bound provides the PAC guarantee through the standard online-to-batch conversion, and its proof is deferred to \cref{app: upper-bound-DMDP}. It is worth noting that \cref{thm: deterministic} only relies on realizability assumption on the Q-function class $\cF$, significantly relaxing the assumptions of realizablity (\cref{asmp:realizable}) and completeness (\cref{asmp:complete}) in \cref{thm: model-free-thm}.

\section{Preference-based Reinforcement Learning}\label{sec: PbRL}

The goal of preference-based learning is to find a near-optimal policy only through interacting with the environment that provides \emph{preference feedback}.
As an extension of our results presented in \cref{sec: upper-bound}, in this section we present a similar algorithm for preference-based RL with the same sample complexity guarantee. %

\newcommand{\pia}[1]{\pi^\iter{#1,+}}
\renewcommand{\pib}[1]{\pi^\iter{#1,-}}
\newcommand{\taua}[1]{\tau^\iter{#1,+}}
\newcommand{\taub}[1]{\tau^\iter{#1,-}}

\newcommand{\Vref}[1]{V^{\rm ref}_{#1}}
\newcommand{\Vhref}[2]{\widehat{V}^{\rm ref}_{#1,#2}}

\paragraph{Preference-based learning in MDP}
In preference-based RL, the interaction protocol of the learner with the environment is specified as follows. For each round $t=1,2,\cdots$,
\begin{itemize}[leftmargin=10pt]
    \item The learner selects policy $\pia{t}$ and $\pib{t}$.
    \item The learner receives trajectories $\taua{t}\sim \pia{t}$, $\taub{t}\sim \pib{t}$, and \emph{preference feedback} $y^\iter{t}\sim \Bern{ \Cmp(\taua{t},\taub{t}) }$, where $\Cmp$ is a comparison function. 
\end{itemize}
Intuitively, for any trajectory pair $(\tau^+,\tau^-)$, the comparison function $\Cmp(\tau^+,\tau^-)=\PP(\tau^+\succ \tau^-)$ measures the probability that $\tau^+$ is more preferred. In this paper, we mainly focus on the Bradley-Terry-Luce (BTR) model~\citep{bradley1952rank}, which is widely used on RLHF literature. We expect that our algorithm and analysis techniques apply to a broader class of preference models. 
\begin{definition}[BTR model]\label{def:BTR}
The comparison function $\Cmp$ is specified as
\begin{align*}
    \Cmp(\tau^+,\tau^-)=\frac{\exp\paren{\beta \Rs(\tau^+)}}{\exp\paren{\beta \Rs(\tau^+)}+\exp\paren{\beta \Rs(\tau^-)}},
\end{align*}
where $\Rs$ is the ground-truth reward function, $\beta>0$ is a parameter.
\end{definition}
Under BTR model, the preference feedback in fact contains information of the outcome rewards. Hence, in this sense, preference-based RL can be regarded as an extension of outcome-based RL with weaker feedback.

\paragraph{Algorithm for preference-based RL}
To extend \cref{alg: model-free}, we need to modify the reward model loss $\LRM$ (defined in \eqref{eq:loss-RM}) to incorporate preference feedback. For any dataset $\cD=\set{(\tau^+,\tau^-,y)}$ consisting of (trajectories, preference) pair, we introduce the following preference-based reward model loss $\LPBRM$:
\begin{align}\label{eq:loss-Pb-RM}
    \LPBRM(R)\defeq \sum_{(\tau^+,\tau^-,y)\in\cD} L\paren{ R(\tau^+)-R(\tau^-), y },
\end{align}
where $L(w,y)\defeq -\beta wy+\log(1+e^{\beta w})$ is the logistic loss. It is well-known that under BTR model (\cref{def:BTR}), the ground-truth reward $\Rs$ is the population minimizer of $\LPBRM$, and any approximate minimizer of $\LPBRM$ can serve as a proxy for $\Rs$. Therefore, with the loss function $\LPBRM$, we propose the following algorithm (\cref{alg: model-free-pbrl}, detailed description in \cref{app: PbRL}), which generalizes \cref{alg: model-free} to handle preference feedback: For each iteration $t=1,2,\cdots,T$, the algorithm performs the following two steps. 
\begin{enumerate}[leftmargin=15pt]
\item (Optimism) Compute optimistic estimates of $(\Qs,\Rs)$ through solving the following joint maximization problem with the dataset $\cD$ consisting of all previously observed (trajectories, feedback) pairs:
\begin{align}\label{eq:max-f-R-pb}
    (f^\iter{t},R^\iter{t})=\max_{f \in \cF, R \in \cR} \lambda \brac{ f_1(s_1) - \Vhref{\cD}{R}} - \LBE(f;R) - \LPBRM(R),
\end{align}
where the Bellman loss $\LBE$ is defined in \eqref{eq:loss-Bellman-full}, $\Vhref{\cD}{R}$ is the estimated value function of $\piexp$ defined as
\begin{align}\label{eq:Vhat-pbrl}
    \Vhref{\cD}{R}\defeq \frac{1}{|\cD|}\sum_{(\tau^+,\tau^-,y)\in\cD} R(\tau^-).
\end{align}
The term $f_1(s_1) - \Vhref{\cD}{R}$ can be regarded as an estimate of the advantage of $\pi_{f}$ over $\piexp$ under $(f,R)$. It is introduced to avoid over-estimating the optimal value, as the preference feedback only provide information between the \emph{difference} between two trajectories. \cfcomment{better explanation needed}

\item (Data collection) The algorithm selects the greedy policy $\pi^\iter{t}\defeq \pi_{f^\iter{t}}$. For each $h\in[H]$, the algorithm sets $\pia{t,h}\defeq \pi\circ_h \piexp$ and $\pib{t,h}\defeq \piexp$, executes $(\pia{t,h},\pib{t,h})$ to collects trajectories $(\taua{t,h}, \taub{t,h})$ and the preference feedback $y^\iter{t,h}$. 
\end{enumerate}
We provide the following sample complexity guarantee of the algorithm above.
\begin{theorem}\label{thm: model-free-thm-pb}
Let $\delta\in(0,1)$. Suppose that \cref{asmp:realizable} and \pref{asmp:complete} hold, and the parameters of \cref{alg: model-free-pbrl} are chosen as
\begin{align}\label{eq: model-free-thm-pb-specs}
    \lambda=c_0\max\sset{ \frac{H\log(\NT{TH^2}/\delta)}{\eps}, TH\epsapp }, \qquad 
    T\geq \tbO{\frac{\ccov H^2}{\eps^2}\cdot \log(\NT{TH^2}/\delta)},
\end{align}
where $c_0>0$ is an absolute constant, and $\tbO{\cdot}$ omits poly-logarithmic factors and constant depending on $\beta$.
Then, \whp, the output policy $\pihat$ of \cref{alg: model-free-pbrl} satisfies
\begin{align*}
    \Vs(s_1)-V^{\pihat}(s_1)\leq \eps+\tbO{\ccov H^2\epsapp}.
\end{align*}
\end{theorem}
The proof of \pref{thm: model-free-thm-pb} is deferred to \pref{sec: proof-preference}.

\section{Lower Bounds}\label{sec: lower-bound}

As shown by \pref{thm: model-free-thm}, with bounded coverability of the MDP and appropriate assumptions on the function classes, finding a near-optimal policy within a polynomial number of episodes with outcome rewards is possible. In this setting, our sample complexity bounds match the sample complexity of Algorithm GOLF \citep{jin2021bellman,xie2022role}, up to a factor of $\tbO{H}$. This indicates that, under bounded coverability, learning with outcome-based rewards is \emph{almost} statistically equivalent to learning with process rewards. 

Additionally, in the setting of offline reinforcement learning with bounded \emph{uniform concentrability}, the results of \citet{jia2025we} indicate that there is also a statistical equivalence between learning with outcome rewards and learning with process rewards. Therefore, it is natural to ask the following question:
\begin{center}
    \emph{Is learning with outcome rewards always statistically equivalent to learning with process rewards in RL?}
\end{center}
However, it turns out that the answer is \emph{negative} if the statistical complexity is measured with respect to the \emph{structure} of the value (reward) function classes. 

More specifically, we construct a class of MDPs with horizon $H=2$, \emph{known} transition $\TT$, and $d$-dimensional \emph{generalized linear} reward models (\cref{app: exp-lower}). With process reward feedback, such a problem is known to be \emph{easy} as it admits low (Bellman) eluder dimension~\citep[etc.]{russo2013eluder,jin2021bellman}, and existing algorithms can learn an $\epsilon$-optimal policy using $\tbO{d^2/\epsilon^2}$ episodes with process rewards. However, given only access to outcome rewards, we show that this problem is \emph{as hard as} learning ReLU linear bandits~\citep{dong2021provable, li2022understanding}, and hence it requires at least $e^{\Omega(d)}$ episodes to learn.
Hence, in this setting, there is an exponential separation between learning with process rewards and learning with outcome-based rewards.
\begin{theorem}\label{thm: lower-bound}
For any positive integer $d\geq 1$, there exists a class $\calM$ of two-layer MDPs with a fixed transition kernel $\TT$ and initial state $s_1$, such that the following holds:
\begin{enumerate}[label=(\alph*)]
\item There exists an algorithm that, for any MDP $\Mstar\in\cM$ and any $\epsilon\in(0,1)$, given access to \emph{process reward} feedback, returns an $\epsilon$-optimal policy with high probability using $\tbO{d^2/\epsilon^2}$ episodes.
\item Suppose that there exists an algorithm that, for any MDP $\Mstar\in\cM$, given only access to \emph{outcome reward}, returns a $0.1$-optimal policy with probability at least $\frac34$ using $T$ episodes. Then it must hold that $T=\Omega(e^{c_1d})$, where $c_1$ is an absolute constant. 
\end{enumerate}
\end{theorem}

This exponential separation demonstrates that the delicate analysis based on well-behaved Bellman errors~\citep[etc.]{jiang2017contextual,jin2021bellman,du2021bilinear} crucially relies on the process reward feedback, and the resulting guarantees might not be preserved in the setting where only outcome reward feedback is available. 

\cfcomment{todo: polish}

\section{Conclusion}
In this work, we develop a model-free, sample-efficient algorithm for outcome-based reinforcement learning that relies solely on trajectory-level rewards and achieves theoretical guarantees bounded by coverability under function approximation. From the lower bound side, we show that joint optimization of reward and value functions is essential, and establish a fundamental exponential gap between outcome-based and per-step feedback. For deterministic MDPs, we propose a simpler, more efficient variant, and extend our approach to preference-based feedback, demonstrating that it preserves the same statistical efficiency.

\section*{Acknowledgements}
We acknowledge support of the Simons Foundation and the NSF through awards DMS-2031883 and PHY-2019786,  ARO through award W911NF-21-1-0328, and the DARPA AIQ award.

\bibliographystyle{plainnat}
\bibliography{ref}

\begin{thebibliography}{54}
\providecommand{\natexlab}[1]{#1}
\providecommand{\url}[1]{\texttt{#1}}
\expandafter\ifx\csname urlstyle\endcsname\relax
  \providecommand{\doi}[1]{doi: #1}\else
  \providecommand{\doi}{doi: \begingroup \urlstyle{rm}\Url}\fi

\bibitem[Amortila et~al.(2024{\natexlab{a}})Amortila, Foster, Jiang, Sekhari, and Xie]{amortila2024harnessing}
Philip Amortila, Dylan~J Foster, Nan Jiang, Ayush Sekhari, and Tengyang Xie.
\newblock Harnessing density ratios for online reinforcement learning.
\newblock \emph{arXiv preprint arXiv:2401.09681}, 2024{\natexlab{a}}.

\bibitem[Amortila et~al.(2024{\natexlab{b}})Amortila, Foster, and Krishnamurthy]{amortila2024scalable}
Philip Amortila, Dylan~J Foster, and Akshay Krishnamurthy.
\newblock Scalable online exploration via coverability.
\newblock \emph{arXiv preprint arXiv:2403.06571}, 2024{\natexlab{b}}.

\bibitem[Antos et~al.(2008)Antos, Szepesv{\'a}ri, and Munos]{antos2008learning}
Andr{\'a}s Antos, Csaba Szepesv{\'a}ri, and R{\'e}mi Munos.
\newblock Learning near-optimal policies with bellman-residual minimization based fitted policy iteration and a single sample path.
\newblock \emph{Machine Learning}, 71:\penalty0 89--129, 2008.

\bibitem[Bai et~al.(2022)Bai, Jones, Ndousse, Askell, Chen, DasSarma, Drain, Fort, Ganguli, Henighan, et~al.]{bai2022training}
Yuntao Bai, Andy Jones, Kamal Ndousse, Amanda Askell, Anna Chen, Nova DasSarma, Dawn Drain, Stanislav Fort, Deep Ganguli, Tom Henighan, et~al.
\newblock Training a helpful and harmless assistant with reinforcement learning from human feedback.
\newblock \emph{arXiv preprint arXiv:2204.05862}, 2022.

\bibitem[Beygelzimer et~al.(2011)Beygelzimer, Langford, Li, Reyzin, and Schapire]{beygelzimer2011contextual}
Alina Beygelzimer, John Langford, Lihong Li, Lev Reyzin, and Robert Schapire.
\newblock Contextual bandit algorithms with supervised learning guarantees.
\newblock In \emph{Proceedings of the Fourteenth International Conference on Artificial Intelligence and Statistics}, pages 19--26. JMLR Workshop and Conference Proceedings, 2011.

\bibitem[Bhardwaj et~al.(2023)Bhardwaj, Xie, Boots, Jiang, and Cheng]{bhardwaj2023adversarial}
Mohak Bhardwaj, Tengyang Xie, Byron Boots, Nan Jiang, and Ching-An Cheng.
\newblock Adversarial model for offline reinforcement learning.
\newblock \emph{Advances in Neural Information Processing Systems}, 36, 2023.

\bibitem[Bradley and Terry(1952)]{bradley1952rank}
Ralph~Allan Bradley and Milton~E Terry.
\newblock Rank analysis of incomplete block designs: I. the method of paired comparisons.
\newblock \emph{Biometrika}, 39\penalty0 (3/4):\penalty0 324--345, 1952.

\bibitem[Cassel et~al.(2024)Cassel, Luo, Rosenberg, and Sotnikov]{cassel2024near}
Asaf Cassel, Haipeng Luo, Aviv Rosenberg, and Dmitry Sotnikov.
\newblock Near-optimal regret in linear mdps with aggregate bandit feedback.
\newblock \emph{arXiv preprint arXiv:2405.07637}, 2024.

\bibitem[Cen et~al.(2024)Cen, Mei, Goshvadi, Dai, Yang, Yang, Schuurmans, Chi, and Dai]{cen2024value}
Shicong Cen, Jincheng Mei, Katayoon Goshvadi, Hanjun Dai, Tong Yang, Sherry Yang, Dale Schuurmans, Yuejie Chi, and Bo~Dai.
\newblock Value-incentivized preference optimization: A unified approach to online and offline rlhf.
\newblock \emph{arXiv preprint arXiv:2405.19320}, 2024.

\bibitem[Chatterji et~al.(2021)Chatterji, Pacchiano, Bartlett, and Jordan]{chatterji2021theory}
Niladri Chatterji, Aldo Pacchiano, Peter Bartlett, and Michael Jordan.
\newblock On the theory of reinforcement learning with once-per-episode feedback.
\newblock \emph{Advances in Neural Information Processing Systems}, 34:\penalty0 3401--3412, 2021.

\bibitem[Chen et~al.(2022{\natexlab{a}})Chen, Mei, and Bai]{chen2022unified}
Fan Chen, Song Mei, and Yu~Bai.
\newblock Unified algorithms for rl with decision-estimation coefficients: pac, reward-free, preference-based learning, and beyond.
\newblock \emph{arXiv preprint arXiv:2209.11745}, 2022{\natexlab{a}}.

\bibitem[Chen et~al.(2024)Chen, Daskalakis, Golowich, and Rakhlin]{chen2024near}
Fan Chen, Constantinos Daskalakis, Noah Golowich, and Alexander Rakhlin.
\newblock Near-optimal learning and planning in separated latent mdps.
\newblock In \emph{The Thirty Seventh Annual Conference on Learning Theory}, pages 995--1067. PMLR, 2024.

\bibitem[Chen and Jiang(2019)]{chen2019information}
Jinglin Chen and Nan Jiang.
\newblock Information-theoretic considerations in batch reinforcement learning.
\newblock In \emph{International Conference on Machine Learning}, pages 1042--1051. PMLR, 2019.

\bibitem[Chen et~al.(2022{\natexlab{b}})Chen, Zhong, Yang, Wang, and Wang]{chen2022human}
Xiaoyu Chen, Han Zhong, Zhuoran Yang, Zhaoran Wang, and Liwei Wang.
\newblock Human-in-the-loop: Provably efficient preference-based reinforcement learning with general function approximation.
\newblock In \emph{International Conference on Machine Learning}, pages 3773--3793. PMLR, 2022{\natexlab{b}}.

\bibitem[Dani et~al.(2008)Dani, Hayes, and Kakade]{dani2008stochastic}
Varsha Dani, Thomas~P Hayes, and Sham~M Kakade.
\newblock Stochastic linear optimization under bandit feedback.
\newblock In \emph{21st Annual Conference on Learning Theory}, number 101, pages 355--366, 2008.

\bibitem[Das et~al.(2024)Das, Chakraborty, Pacchiano, and Chowdhury]{das2024provably}
Nirjhar Das, Souradip Chakraborty, Aldo Pacchiano, and Sayak~Ray Chowdhury.
\newblock Provably sample efficient rlhf via active preference optimization.
\newblock \emph{arXiv preprint arXiv:2402.10500}, 2024.

\bibitem[Dong et~al.(2021)Dong, Yang, and Ma]{dong2021provable}
Kefan Dong, Jiaqi Yang, and Tengyu Ma.
\newblock Provable model-based nonlinear bandit and reinforcement learning: Shelve optimism, embrace virtual curvature.
\newblock \emph{Advances in Neural Information Processing Systems}, 34:\penalty0 26168--26182, 2021.

\bibitem[Du et~al.(2021)Du, Kakade, Lee, Lovett, Mahajan, Sun, and Wang]{du2021bilinear}
Simon Du, Sham Kakade, Jason Lee, Shachar Lovett, Gaurav Mahajan, Wen Sun, and Ruosong Wang.
\newblock Bilinear classes: A structural framework for provable generalization in rl.
\newblock In \emph{International Conference on Machine Learning}, pages 2826--2836. PMLR, 2021.

\bibitem[Efroni et~al.(2021)Efroni, Merlis, and Mannor]{efroni2021reinforcement}
Yonathan Efroni, Nadav Merlis, and Shie Mannor.
\newblock Reinforcement learning with trajectory feedback.
\newblock In \emph{Proceedings of the AAAI conference on artificial intelligence}, volume~35, pages 7288--7295, 2021.

\bibitem[Farahmand et~al.(2010)Farahmand, Szepesv{\'a}ri, and Munos]{farahmand2010error}
Amir-massoud Farahmand, Csaba Szepesv{\'a}ri, and R{\'e}mi Munos.
\newblock Error propagation for approximate policy and value iteration.
\newblock \emph{Advances in neural information processing systems}, 23, 2010.

\bibitem[Foster et~al.(2021)Foster, Kakade, Qian, and Rakhlin]{foster2021statistical}
Dylan~J Foster, Sham~M Kakade, Jian Qian, and Alexander Rakhlin.
\newblock The statistical complexity of interactive decision making.
\newblock \emph{arXiv preprint arXiv:2112.13487}, 2021.

\bibitem[Foster et~al.(2022)Foster, Rakhlin, Sekhari, and Sridharan]{foster2022complexity}
Dylan~J Foster, Alexander Rakhlin, Ayush Sekhari, and Karthik Sridharan.
\newblock On the complexity of adversarial decision making.
\newblock \emph{Advances in Neural Information Processing Systems}, 35:\penalty0 35404--35417, 2022.

\bibitem[Jaech et~al.(2024)Jaech, Kalai, Lerer, Richardson, El-Kishky, Low, Helyar, Madry, Beutel, Carney, et~al.]{jaech2024openai}
Aaron Jaech, Adam Kalai, Adam Lerer, Adam Richardson, Ahmed El-Kishky, Aiden Low, Alec Helyar, Aleksander Madry, Alex Beutel, Alex Carney, et~al.
\newblock Openai o1 system card.
\newblock \emph{arXiv preprint arXiv:2412.16720}, 2024.

\bibitem[Jia et~al.(2025)Jia, Rakhlin, and Xie]{jia2025we}
Zeyu Jia, Alexander Rakhlin, and Tengyang Xie.
\newblock Do we need to verify step by step? rethinking process supervision from a theoretical perspective.
\newblock \emph{arXiv preprint arXiv:2502.10581}, 2025.

\bibitem[Jiang et~al.(2017)Jiang, Krishnamurthy, Agarwal, Langford, and Schapire]{jiang2017contextual}
Nan Jiang, Akshay Krishnamurthy, Alekh Agarwal, John Langford, and Robert~E Schapire.
\newblock Contextual decision processes with low bellman rank are pac-learnable.
\newblock In \emph{International Conference on Machine Learning}, pages 1704--1713. PMLR, 2017.

\bibitem[Jin et~al.(2021{\natexlab{a}})Jin, Liu, and Miryoosefi]{jin2021bellman}
Chi Jin, Qinghua Liu, and Sobhan Miryoosefi.
\newblock Bellman eluder dimension: New rich classes of rl problems, and sample-efficient algorithms.
\newblock \emph{Advances in Neural Information Processing Systems}, 34:\penalty0 13406--13418, 2021{\natexlab{a}}.

\bibitem[Jin et~al.(2021{\natexlab{b}})Jin, Yang, and Wang]{jin2021pessimism}
Ying Jin, Zhuoran Yang, and Zhaoran Wang.
\newblock Is pessimism provably efficient for offline rl?
\newblock In \emph{International Conference on Machine Learning}, pages 5084--5096. PMLR, 2021{\natexlab{b}}.

\bibitem[Kakade and Langford(2002)]{kakade2002approximately}
Sham Kakade and John Langford.
\newblock Approximately optimal approximate reinforcement learning.
\newblock In \emph{ICML}, volume~2, pages 267--274, 2002.

\bibitem[Lancewicki and Mansour(2025)]{lancewicki2025near}
Tal Lancewicki and Yishay Mansour.
\newblock Near-optimal regret using policy optimization in online mdps with aggregate bandit feedback.
\newblock \emph{arXiv preprint arXiv:2502.04004}, 2025.

\bibitem[Lattimore and Szepesv{\'a}ri(2020)]{lattimore2020bandit}
Tor Lattimore and Csaba Szepesv{\'a}ri.
\newblock \emph{Bandit algorithms}.
\newblock Cambridge University Press, 2020.

\bibitem[Li et~al.(2022)Li, Kamath, Foster, and Srebro]{li2022understanding}
Gene Li, Pritish Kamath, Dylan~J Foster, and Nati Srebro.
\newblock Understanding the eluder dimension.
\newblock \emph{Advances in Neural Information Processing Systems}, 35:\penalty0 23737--23750, 2022.

\bibitem[Liu et~al.(2023)Liu, Viano, and Cevher]{liu2023can}
Fanghui Liu, Luca Viano, and Volkan Cevher.
\newblock What can online reinforcement learning with function approximation benefit from general coverage conditions?
\newblock In \emph{International Conference on Machine Learning}, pages 22063--22091. PMLR, 2023.

\bibitem[Munos(2003)]{munos2003error}
R{\'e}mi Munos.
\newblock Error bounds for approximate policy iteration.
\newblock In \emph{ICML}, volume~3, pages 560--567. Citeseer, 2003.

\bibitem[Neu and Bart{\'o}k(2013)]{neu2013efficient}
Gergely Neu and G{\'a}bor Bart{\'o}k.
\newblock An efficient algorithm for learning with semi-bandit feedback.
\newblock In \emph{International Conference on Algorithmic Learning Theory}, pages 234--248. Springer, 2013.

\bibitem[Novoseller et~al.(2020)Novoseller, Wei, Sui, Yue, and Burdick]{novoseller2020dueling}
Ellen Novoseller, Yibing Wei, Yanan Sui, Yisong Yue, and Joel Burdick.
\newblock Dueling posterior sampling for preference-based reinforcement learning.
\newblock In \emph{Conference on Uncertainty in Artificial Intelligence}, pages 1029--1038. PMLR, 2020.

\bibitem[Osband and Van~Roy(2014)]{osband2014model}
Ian Osband and Benjamin Van~Roy.
\newblock Model-based reinforcement learning and the eluder dimension.
\newblock In \emph{Advances in Neural Information Processing Systems}, volume~27, pages 1466--1474, 2014.

\bibitem[Ouyang et~al.(2022)Ouyang, Wu, Jiang, Almeida, Wainwright, Mishkin, Zhang, Agarwal, Slama, Ray, et~al.]{ouyang2022training}
Long Ouyang, Jeffrey Wu, Xu~Jiang, Diogo Almeida, Carroll Wainwright, Pamela Mishkin, Chong Zhang, Sandhini Agarwal, Katarina Slama, Alex Ray, et~al.
\newblock Training language models to follow instructions with human feedback.
\newblock \emph{Advances in Neural Information Processing Systems}, 35:\penalty0 27730--27744, 2022.

\bibitem[Pacchiano et~al.(2021)Pacchiano, Saha, and Lee]{pacchiano2021dueling}
Aldo Pacchiano, Aadirupa Saha, and Jonathan Lee.
\newblock Dueling rl: reinforcement learning with trajectory preferences.
\newblock \emph{arXiv preprint arXiv:2111.04850}, 2021.

\bibitem[Russo and Van~Roy(2013)]{russo2013eluder}
Daniel Russo and Benjamin Van~Roy.
\newblock Eluder dimension and the sample complexity of optimistic exploration.
\newblock In \emph{Advances in Neural Information Processing Systems}, volume~26, pages 2256--2264, 2013.

\bibitem[Sun et~al.(2019)Sun, Jiang, Krishnamurthy, Agarwal, and Langford]{sun2019model}
Wen Sun, Nan Jiang, Akshay Krishnamurthy, Alekh Agarwal, and John Langford.
\newblock Model-based {RL} in contextual decision processes: {PAC} bounds and exponential improvements over model-free approaches.
\newblock In \emph{Conference on learning theory}, pages 2898--2933. PMLR, 2019.

\bibitem[Sutton et~al.(1998)Sutton, Barto, et~al.]{sutton1998reinforcement}
Richard~S Sutton, Andrew~G Barto, et~al.
\newblock \emph{Reinforcement learning: An introduction}, volume~1.
\newblock MIT press Cambridge, 1998.

\bibitem[Wang et~al.(2023)Wang, Liu, and Jin]{wang2023rlhf}
Yuanhao Wang, Qinghua Liu, and Chi Jin.
\newblock Is rlhf more difficult than standard rl?
\newblock \emph{arXiv preprint arXiv:2306.14111}, 2023.

\bibitem[Wu and Sun(2023)]{wu2023making}
Runzhe Wu and Wen Sun.
\newblock Making rl with preference-based feedback efficient via randomization.
\newblock \emph{arXiv preprint arXiv:2310.14554}, 2023.

\bibitem[Xie and Jiang(2021)]{xie2021batch}
Tengyang Xie and Nan Jiang.
\newblock Batch value-function approximation with only realizability.
\newblock In \emph{International Conference on Machine Learning}, pages 11404--11413. PMLR, 2021.

\bibitem[Xie et~al.(2021)Xie, Cheng, Jiang, Mineiro, and Agarwal]{xie2021bellman}
Tengyang Xie, Ching-An Cheng, Nan Jiang, Paul Mineiro, and Alekh Agarwal.
\newblock Bellman-consistent pessimism for offline reinforcement learning.
\newblock \emph{Advances in neural information processing systems}, 34:\penalty0 6683--6694, 2021.

\bibitem[Xie et~al.(2022)Xie, Foster, Bai, Jiang, and Kakade]{xie2022role}
Tengyang Xie, Dylan~J Foster, Yu~Bai, Nan Jiang, and Sham~M Kakade.
\newblock The role of coverage in online reinforcement learning.
\newblock \emph{arXiv preprint arXiv:2210.04157}, 2022.

\bibitem[Xie et~al.(2024)Xie, Foster, Krishnamurthy, Rosset, Awadallah, and Rakhlin]{xie2024exploratory}
Tengyang Xie, Dylan~J Foster, Akshay Krishnamurthy, Corby Rosset, Ahmed Awadallah, and Alexander Rakhlin.
\newblock Exploratory preference optimization: Harnessing implicit q*-approximation for sample-efficient rlhf.
\newblock \emph{arXiv preprint arXiv:2405.21046}, 2024.

\bibitem[Xu et~al.(2020)Xu, Wang, Yang, Singh, and Dubrawski]{xu2020preference}
Yichong Xu, Ruosong Wang, Lin Yang, Aarti Singh, and Artur Dubrawski.
\newblock Preference-based reinforcement learning with finite-time guarantees.
\newblock \emph{Advances in Neural Information Processing Systems}, 33:\penalty0 18784--18794, 2020.

\bibitem[Ye et~al.(2024)Ye, Xiong, Zhang, Jiang, and Zhang]{ye2024theoretical}
Chenlu Ye, Wei Xiong, Yuheng Zhang, Nan Jiang, and Tong Zhang.
\newblock A theoretical analysis of nash learning from human feedback under general kl-regularized preference.
\newblock \emph{arXiv preprint arXiv:2402.07314}, 2024.

\bibitem[Zanette et~al.(2020)Zanette, Lazaric, Kochenderfer, and Brunskill]{zanette2020learning}
Andrea Zanette, Alessandro Lazaric, Mykel Kochenderfer, and Emma Brunskill.
\newblock Learning near optimal policies with low inherent bellman error.
\newblock In \emph{International Conference on Machine Learning}, pages 10978--10989. PMLR, 2020.

\bibitem[Zhan et~al.(2023)Zhan, Uehara, Kallus, Lee, and Sun]{zhan2023provable}
Wenhao Zhan, Masatoshi Uehara, Nathan Kallus, Jason~D Lee, and Wen Sun.
\newblock Provable offline preference-based reinforcement learning.
\newblock \emph{arXiv preprint arXiv:2305.14816}, 2023.

\bibitem[Zhang et~al.(2024)Zhang, Yu, Sharma, Zhong, Liu, Yang, Wang, Hassan, and Wang]{zhang2024self}
Shenao Zhang, Donghan Yu, Hiteshi Sharma, Han Zhong, Zhihan Liu, Ziyi Yang, Shuohang Wang, Hany Hassan, and Zhaoran Wang.
\newblock Self-exploring language models: Active preference elicitation for online alignment.
\newblock \emph{arXiv preprint arXiv:2405.19332}, 2024.

\bibitem[Zhang(2002)]{zhang2002covering}
Tong Zhang.
\newblock Covering number bounds of certain regularized linear function classes.
\newblock \emph{Journal of Machine Learning Research}, 2\penalty0 (Mar):\penalty0 527--550, 2002.

\bibitem[Zhu et~al.(2023)Zhu, Jordan, and Jiao]{zhu2023principled}
Banghua Zhu, Michael Jordan, and Jiantao Jiao.
\newblock Principled reinforcement learning with human feedback from pairwise or k-wise comparisons.
\newblock In \emph{International Conference on Machine Learning}, pages 43037--43067. PMLR, 2023.

\end{thebibliography}

\neurips{
\clearpage

\clearpage
\section*{NeurIPS Paper Checklist}

The checklist is designed to encourage best practices for responsible machine learning research, addressing issues of reproducibility, transparency, research ethics, and societal impact. Do not remove the checklist: {\bf The papers not including the checklist will be desk rejected.} The checklist should follow the references and follow the (optional) supplemental material.  The checklist does NOT count towards the page
limit. 

Please read the checklist guidelines carefully for information on how to answer these questions. For each question in the checklist:
\begin{itemize}
    \item You should answer \answerYes{}, \answerNo{}, or \answerNA{}.
    \item \answerNA{} means either that the question is Not Applicable for that particular paper or the relevant information is Not Available.
    \item Please provide a short (1–2 sentence) justification right after your answer (even for NA). 
\end{itemize}

{\bf The checklist answers are an integral part of your paper submission.} They are visible to the reviewers, area chairs, senior area chairs, and ethics reviewers. You will be asked to also include it (after eventual revisions) with the final version of your paper, and its final version will be published with the paper.

The reviewers of your paper will be asked to use the checklist as one of the factors in their evaluation. While "\answerYes{}" is generally preferable to "\answerNo{}", it is perfectly acceptable to answer "\answerNo{}" provided a proper justification is given (e.g., "error bars are not reported because it would be too computationally expensive" or "we were unable to find the license for the dataset we used"). In general, answering "\answerNo{}" or "\answerNA{}" is not grounds for rejection. While the questions are phrased in a binary way, we acknowledge that the true answer is often more nuanced, so please just use your best judgment and write a justification to elaborate. All supporting evidence can appear either in the main paper or the supplemental material, provided in appendix. If you answer \answerYes{} to a question, in the justification please point to the section(s) where related material for the question can be found.

IMPORTANT, please:
\begin{itemize}
    \item {\bf Delete this instruction block, but keep the section heading ``NeurIPS Paper Checklist"},
    \item  {\bf Keep the checklist subsection headings, questions/answers and guidelines below.}
    \item {\bf Do not modify the questions and only use the provided macros for your answers}.
\end{itemize}

\begin{enumerate}

\item {\bf Claims}
    \item[] Question: Do the main claims made in the abstract and introduction accurately reflect the paper's contributions and scope?
    \item[] Answer: \answerYes{} %
    \item[] Justification: Claims made in the abstract and introduction accurately reflect this paper’s contributions and scope. More supporting details are included in the main text.

\item {\bf Limitations}
    \item[] Question: Does the paper discuss the limitations of the work performed by the authors?
    \item[] Answer: \answerYes{} %
    \item[] Justification: This paper discussed the limitations in the main body below each theorem.

\item {\bf Theory assumptions and proofs}
    \item[] Question: For each theoretical result, does the paper provide the full set of assumptions and a complete (and correct) proof?
    \item[] Answer: \answerYes{} %
    \item[] Justification: This paper provides the full set of assumptions in the main body and a complete (and correct) proof in the appendix.

    \item {\bf Experimental result reproducibility}
    \item[] Question: Does the paper fully disclose all the information needed to reproduce the main experimental results of the paper to the extent that it affects the main claims and/or conclusions of the paper (regardless of whether the code and data are provided or not)?
    \item[] Answer: \answerNA{} %
    \item[] Justification: This paper does not include experiments.

\item {\bf Open access to data and code}
    \item[] Question: Does the paper provide open access to the data and code, with sufficient instructions to faithfully reproduce the main experimental results, as described in supplemental material?
    \item[] Answer: \answerNA{} %
    \item[] Justification: This paper does not include experiments requiring code.

\item {\bf Experimental setting/details}
    \item[] Question: Does the paper specify all the training and test details (e.g., data splits, hyperparameters, how they were chosen, type of optimizer, etc.) necessary to understand the results?
    \item[] Answer: \answerNA{} %
    \item[] Justification: This paper does not include experiments.

\item {\bf Experiment statistical significance}
    \item[] Question: Does the paper report error bars suitably and correctly defined or other appropriate information about the statistical significance of the experiments?
    \item[] Answer: \answerNA{} %
    \item[] Justification: This paper does not include experiments.

\item {\bf Experiments compute resources}
    \item[] Question: For each experiment, does the paper provide sufficient information on the computer resources (type of compute workers, memory, time of execution) needed to reproduce the experiments?
    \item[] Answer: \answerNA{} %
    \item[] Justification: The paper does not include experiments.
    
\item {\bf Code of ethics}
    \item[] Question: Does the research conducted in the paper conform, in every respect, with the NeurIPS Code of Ethics \url{https://neurips.cc/public/EthicsGuidelines}?
    \item[] Answer: \answerYes{} %
    \item[] Justification: The authors have reviewed the NeurIPS Code of Ethics.

\item {\bf Broader impacts}
    \item[] Question: Does the paper discuss both potential positive societal impacts and negative societal impacts of the work performed?
    \item[] Answer: \answerNA{} %
    \item[] Justification: This is a pure theoretical paper. There is no societal impact of the work performed.
    
\item {\bf Safeguards}
    \item[] Question: Does the paper describe safeguards that have been put in place for responsible release of data or models that have a high risk for misuse (e.g., pretrained language models, image generators, or scraped datasets)?
    \item[] Answer: \answerNA{} %
    \item[] Justification: This paper poses no such risks.

\item {\bf Licenses for existing assets}
    \item[] Question: Are the creators or original owners of assets (e.g., code, data, models), used in the paper, properly credited and are the license and terms of use explicitly mentioned and properly respected?
    \item[] Answer: \answerNA{} %
    \item[] Justification: This paper does not use existing assets.

\item {\bf New assets}
    \item[] Question: Are new assets introduced in the paper well documented and is the documentation provided alongside the assets?
    \item[] Answer: \answerNA{} %
    \item[] Justification: This paper does not release new assets.

\item {\bf Crowdsourcing and research with human subjects}
    \item[] Question: For crowdsourcing experiments and research with human subjects, does the paper include the full text of instructions given to participants and screenshots, if applicable, as well as details about compensation (if any)? 
    \item[] Answer: \answerNA{} %
    \item[] Justification: This paper does not involve crowdsourcing and research with human subjects.

\item {\bf Institutional review board (IRB) approvals or equivalent for research with human subjects}
    \item[] Question: Does the paper describe potential risks incurred by study participants, whether such risks were disclosed to the subjects, and whether Institutional Review Board (IRB) approvals (or an equivalent approval/review based on the requirements of your country or institution) were obtained?
    \item[] Answer: \answerNA{} %
    \item[] Justification: This paper does not involve research with human subjects.

\item {\bf Declaration of LLM usage}
    \item[] Question: Does the paper describe the usage of LLMs if it is an important, original, or non-standard component of the core methods in this research? Note that if the LLM is used only for writing, editing, or formatting purposes and does not impact the core methodology, scientific rigorousness, or originality of the research, declaration is not required.
    \item[] Answer: \answerNA{} %
    \item[] Justification: The core method development in this research does not involve LLMs as any important, original, or non-standard components.

\end{enumerate}

}

\newpage

\appendix
\allowdisplaybreaks

\section{More Related Works}
We review more related works in this section.

The \emph{coverability coefficient} has recently gained attention in the theory of online reinforcement learning \citep{xie2022role,liu2023can,amortila2024harnessing,amortila2024scalable}. This condition is in the same spirit as the widely used concentrability coefficient \citep{munos2003error,antos2008learning,farahmand2010error,chen2019information,jin2021pessimism,xie2021batch,xie2021bellman,bhardwaj2023adversarial}, a concept frequently employed in the theory of offline (or batch) reinforcement learning.  A well-known duality suggests that the coverability coefficient can be interpreted as the optimal concentrability coefficient attainable by any offline data distribution. For further discussion, see \cite{xie2022role}.

A related body of theoretical work explores \emph{reinforcement learning with trajectory feedback} \citep{neu2013efficient,efroni2021reinforcement,chatterji2021theory,cassel2024near,lancewicki2025near}, where the learner receives only episode-level feedback at the end of each trajectory. This category also encompasses preference-based reinforcement learning \citep{pacchiano2021dueling,chen2022human,zhu2023principled,wu2023making,zhan2023provable}, which relies on pairwise comparisons between trajectories. While most prior work focuses on tabular or linear MDP settings, we take a step further by studying learning with function approximation, and bound the complexity by the coverability coefficient.
\cfcomment{Trajectory eluder dimension suffers the same lower bound \& CB}

In the context of \emph{Online Reinforcement Learning}, numerous prior works have investigated the complexity of exploration and policy optimization, introducing various complexity measures such as Bellman rank \citep{jiang2017contextual}, Eluder dimension \citep{russo2013eluder, osband2014model}, witness rank \citep{sun2019model}, Bellman-Eluder dimension \citep{jin2021bellman}, the bilinear class \citep{du2021bilinear}, and decision-estimation coefficients \citep{foster2021statistical}. These complexity notions characterize properties of the function or model class but are generally not instance-dependent. In contrast, \citet{xie2022role} introduces the instance-dependent notion of \emph{coverability coefficients} to provide complexity bounds in online reinforcement learning. For further discussion on instance-dependent complexity measures, we refer the reader to the discussions therein.

We further review some literatures on \emph{online preference-based learning} or \emph{online RLHF}. \cite{xu2020preference, novoseller2020dueling, pacchiano2021dueling, wu2023making, zhan2023provable, das2024provably} provides theoretical guarantees for tabular MDPs and linear MDPs. \cite{ye2024theoretical} studies RLHF with general function approximation for contextual bandits, which is equivalent to the case where $H = 1$. \cite{chen2022human, wang2023rlhf} use the Eluder dimension type complexity measures to characterize the sample complexity of online RLHF, which sometimes can be too pessimistic. \cite{xie2024exploratory, cen2024value, zhang2024self} proposed algorithms for online RLHF with function approximation, but their complexity depends on the trajectory coverability instead of the state coverability

\section{Technical tools}

\subsection{Uniform convergence with square loss}

\newcommand{\hs}{F^\star}
\newcommand{\hsp}{F^\sharp}

To prove the uniform convergence results with square loss, we frequently use the following version Freedman's inequality \citep[see e.g.,][]{beygelzimer2011contextual}.
\begin{lemma}[Freedman's inequality]\label{lem:freedman}
Suppose that $Z\ind{1},\cdots,Z\ind{T}$ is a martingale difference sequence that is adapted to the filtration $(\filt\ind{t})_{t=1}^T$, and $Z\ind{t}\leq C$ almost surely for all $t\in[T]$. Then for any $\lambda\in[0,\frac1C]$, \whp, for all $n\leq T$,
\begin{align*}
    \sum_{t=1}^n Z\ind{t}\leq \lambda \sum_{t=1}^n \EE\cond{(Z\ind{t})^2}{\filt\ind{t-1}}+\frac{\log(1/\delta)}{\lambda}.
\end{align*}
\end{lemma}

\begin{lemma}\label{lem:ERM-single}
Suppose that $(x\ind{1},y\ind{1}),\cdots,(x\ind{T},y\ind{T})$ is a sequence of random variable in $\cX\times[0,C]$ that is adapted to the filtration $(\filt\ind{t})_{t=1}^T$, %
such that there exists a function $\hs:\cX\to[0,1]$ with $\hs(x\ind{t})=\EE[y\ind{t}|\filt\ind{t-1},x\ind{t}]$ almost surely. Then for any function $F:\cX\to[0,C]$, it holds that \whp, for all $n\in[T]$,
\begin{align*}
    \sum_{t=1}^n \paren{F(x\ind{t})-y\ind{t}}^2- \sum_{t=1}^n \paren{\hs(x\ind{t})-y\ind{t}}^2\geq \frac{1}{2}\sum_{t=1}^n \EE\cond{\paren{F(x\ind{t})-\hs(x\ind{t})}^2}{\filt\ind{t-1}}-10C^2\log(1/\delta).
\end{align*}
Conversely, it holds that \whp, for all $n\in[T]$,
\begin{align*}
    \sum_{t=1}^n \paren{F(x\ind{t})-y\ind{t}}^2- \sum_{t=1}^n \paren{\hs(x\ind{t})-y\ind{t}}^2\leq 2\sum_{t=1}^n \EE\cond{\paren{F(x\ind{t})-\hs(x\ind{t})}^2}{\filt\ind{t-1}}+5C^2\log(1/\delta).
\end{align*}
\end{lemma}

\begin{proofof}{\cref{lem:ERM-single}}
Denote
\begin{align*}
    W\ind{t}\defeq&~ \paren{F(x\ind{t})-y\ind{t}}^2- \paren{\hs(x\ind{t})-y\ind{t}}^2 \\
    =&~ \paren{F(x\ind{t})-\hs(x\ind{t})}^2+2\paren{F(x\ind{t})-\hs(x\ind{t})}\paren{\hs(x\ind{t})-y\ind{t}}.
\end{align*}
Note that
\begin{align*}
    \EE\cond{W\ind{t}}{\filt\ind{t-1}}=\EE\cond{\paren{F(x\ind{t})-\hs(x\ind{t})}^2}{\filt\ind{t-1}},
\end{align*}
and
\begin{align*}
    Z\ind{t}\defeq W\ind{t}-\EE\cond{W\ind{t}}{\filt\ind{t-1}}\leq W\ind{t}\leq C^2.
\end{align*}
Therefore, using Freedman's inequality (\cref{lem:freedman}), for any fixed $\lambda\in[0,\frac1{C^2}]$, we have \whp, 
\begin{align*}
    \sum_{t=1}^n Z\ind{t}\leq \lambda \sum_{t=1}^n \EE\cond{(Z\ind{t})^2}{\filt\ind{t-1}}+\frac{\log(1/\delta)}{\lambda}, \qquad \forall n\in[T].
\end{align*}
Note that
\begin{align*}
    \EE\cond{(Z\ind{t})^2}{\filt\ind{t-1}}\leq&~ \EE\cond{(W\ind{t})^2}{\filt\ind{t-1}} \\
    =&~ \EE\cond{\paren{F(x\ind{t})-\hs(x\ind{t})}^4+4\paren{F(x\ind{t})-\hs(x\ind{t})}^2\paren{\hs(x\ind{t})-y\ind{t}}^2}{\filt\ind{t-1}} \\
    \leq&~ 5C^2 \EE\cond{\paren{F(x\ind{t})-\hs(x\ind{t})}^2}{\filt\ind{t-1}}.
\end{align*}
Therefore, by setting $\lambda=\frac{1}{5C^2}$, we get the desired upper bound. 

Similarly, for the lower bound, we can apply Freedman's inequality with $(-Z\ind{t})$ to show that for $\lambda=\frac{1}{10C^2}$, \whp,
\begin{align*}
    -\sum_{t=1}^n Z\ind{t}\leq&~ \lambda \sum_{t=1}^n \EE\cond{(Z\ind{t})^2}{\filt\ind{t-1}}+\frac{\log(1/\delta)}{\lambda} \\
    \leq&~ \frac{1}{2}\sum_{t=1}^n \EE\cond{\paren{F(x\ind{t})-\hs(x\ind{t})}^2}{\filt\ind{t-1}}+10C^2\log(1/\delta), \qquad \forall n\in[T].
\end{align*}
\end{proofof}

\begin{proposition}\label{lem:ERM-unif}
Fix a parameter $\alpha\geq 0$. Under the assumption of \cref{lem:ERM-single}, suppose that $\cH\subseteq (\cX\to [0,C])$ is a fixed function class, and $\hsp\in\cH$ satisfies $\linf{\hs-\hsp}\leq \epsapp$. Define
\begin{align*}
    \cL_n(F)\defeq \sum_{t=1}^n \paren{F(x\ind{t})-y\ind{t}}^2, \qquad
    \cE_n(F)\defeq \sum_{t=1}^n \EE\cond{\paren{F(x\ind{t})-\hs(x\ind{t})}^2}{\filt\ind{t-1}}.
\end{align*}
Let $\kappa\defeq 15C^2\log(2N(\cH,\alpha)/\delta)+3Cn\alpha+4n\epsapp^2$.
Then the following holds simultaneously \whp:

(1) For each $n\in[T]$,
\begin{align*}
    \cL_n(\hsp)-\inf_{F'\in\cH}\cL_n(F')\leq \kappa.
\end{align*}

(2) For each $n\in[T]$, for all $F\in\cH$, %
\begin{align*}
    \frac12 \cE_n(F)\leq \cL_n(F)-\inf_{F'\in\cH}\cL_n(F')+\kappa.
\end{align*}
\end{proposition}

\begin{proofof}{\cref{lem:ERM-unif}}
Denote $N\defeq N(\cH,\alpha)$. Let $\cH_\alpha$ be a minimal $\alpha$-covering of $\cH$. Then applying \cref{lem:ERM-single} and the union bound, we have \whp, the following holds simultaneously for $n\in[T]$:

(1) For all $F'\in\cH_\alpha$, %
it holds that
\begin{align*}
    \frac12\cE_n(F')\leq \cL_n(F')-\cL_n(\hs)+10C^2\log(2N/\delta). 
\end{align*}
(2) It holds that
\begin{align*}
    \cL_n(\hsp)-\cL_n(\hs)\leq 2\cE_n(\hsp)+5C^2\log(2/\delta).
\end{align*}
In the following, we condition on the above success event.

By definition, $\cE_n(\hsp)\leq n\epsapp^2$, and hence
\begin{align}\label{eq:proof-ERM-unif-1}
    \cL_n(\hs)\geq \cL_n(\hsp)-4n\epsapp^2-5C^2\log(2/\delta).
\end{align}
Furthermore, for any $F\in\cH$, there exists $F'\in\cH_\alpha$ with $\linf{F-F'}\leq \alpha$, which implies
\begin{align*}
    \abs{\cL_n(F)-\cL_n(F')}\leq 2Cn\alpha, \qquad
    \abs{\cE_n(F)-\cE_n(F')}\leq 2Cn\alpha.
\end{align*}
Therefore, under the success event, we have
\begin{align*}
    \frac12\cE_n(F)\leq \cL_n(F)-\cL_n(\hs)+10C^2\log(2N/\delta)+3Cn\alpha.
\end{align*}
holds for arbitrary $F\in\cF$. Hence, by \eqref{eq:proof-ERM-unif-1}, we have
\begin{align*}
    \frac12\cE_n(F)\leq \cL_n(F)-\cL_n(\hsp)+15C^2\log(2N/\delta)+3Cn\alpha+4n\epsapp^2.
\end{align*}
Noting that $\cL_n(\hsp)\geq \inf_{F'\in\cH} \cL_n(F')$ completes the proof of (2).
Furthermore, using $\cE_n(F)\geq 0$, we also have %
\begin{align*}
    \cL_n(\hsp)\leq \inf_{F'\in\cH} \cL_n(F')+15C^2\log(2N/\delta)+3Cn\alpha+4n\epsapp^2.
\end{align*}
This completes the proof of (1).
\end{proofof}

\subsection{Uniform convergence with log-loss}

We prove the following result, which is a direct extension of the standard MLE guarantee~\citep{zhang2002covering}.

\newcommand{\Nlog}{N_{\log}}
\newcommand{\ths}{{\theta^\star}}
\newcommand{\DHs}[1]{D_{\rm H}^2\paren{#1}}

\begin{proposition}\label{prop:MLE-unif}
Suppose that $\set{P_\theta(y|x)}_{\theta\in\Theta}\subseteq (\cX\to\Delta(\cY))$ is a class of condition densities parametrized by an abstract parameter class $\Theta$. Without loss of generality, we assume $\cY$ is discrete.

A $\alpha$-covering of $\Theta$ is a subset $\Theta'\subseteq \Theta$ such that for any $\theta\in\Theta$, there exists $\theta'\in\Theta'$ such that $\abs{\log P_\theta(y|x)-\log P_{\theta'}(y|x)}\leq \alpha$ for all $x\in\cX, y\in\cY$. 
We define the covering number of $\Theta$ under log-loss as
\begin{align*}
    \Nlog(\Theta,\alpha)\defeq \min\set{|\Theta'|: \Theta' \text{ is a $\alpha$-covering of }\Theta}.
\end{align*}

Suppose that $(x\ind{1},y\ind{1}),\cdots,(x\ind{T},y\ind{T})$ is a sequence of random variables adapted to the filtration $(\filt\ind{t})_{t=1}^T$, such that there exists $\ths\in\Theta$ so that $\PP(y\ind{t}=\cdot|x\ind{t},\filt\ind{t-1})= P_{\ths}(y\ind{t}=\cdot|x\ind{t})$ almost surely for $t\in[T]$. Then it holds that for all $n\in[T]$, for all $\theta\in\Theta$,
\begin{align*}
    \sum_{t=1}^n \EE\cond{ \DHs{ P_\theta(\cdot|x\ind{t}), P_\ths(\cdot|x\ind{t}) } }{\filt\ind{t-1}}
    \leq&~ -\frac12 \sum_{t=1}^n \brac{ \log P_\theta(y\ind{t}|x\ind{t})-\log P_\ths(y\ind{t}|x\ind{t}) } \\
    &~ + \log \Nlog(\Theta,\alpha)+2n\alpha.
\end{align*}
\end{proposition}

\begin{proofof}{\cref{prop:MLE-unif}}
Let $\Theta'\subseteq \Theta$ be a minimal $\alpha$-covering, and let $N\defeq |\Theta'|=\Nlog(\Theta,\alpha)$. For each $\theta\in\Theta$, we consider
\begin{align*}
    L\ind{t}(\theta)\defeq \log P_\theta(y\ind{t}|x\ind{t})-\log P_\ths(y\ind{t}|x\ind{t}).
\end{align*}
Then it holds that
\begin{align*}
    \EE\cond{ \exp\paren{ \frac{1}{2} L\ind{t}(\theta) } }{x\ind{t}, \filt\ind{t-1}}
    =&~\EE_{y\sim P_\ths(\cdot|x\ind{t})} \sqrt{\frac{P_\theta(y|x\ind{t})}{P_\ths(y|x\ind{t})}} \\
    =&~\sum_{y\in\cY} \sqrt{P_\theta(y|x\ind{t}) P_\ths(y|x\ind{t})} \\
    =&~ 1- \DHs{ P_\theta(\cdot|x\ind{t}), P_\ths(\cdot|x\ind{t}) }.
\end{align*}
Therefore, applying \cref{lemma:concen} and using union bound over $\theta\in\Theta'$, we have the following bound: \whp, for any $\theta'\in\Theta'$, $n\in[T]$,
\begin{align*}
    \sum_{t=1}^{n} -\log \cond{\exp\paren{ \frac12 L\ind{t}(\theta')}}{\cF\ind{t-1}} \leq - \frac12 \sum_{t=1}^{n} L\ind{t}(\theta') +\log(N/\delta).
\end{align*}
In the following, we condition on the above event. Fix any $\theta\in\Theta$. Then, there exists $\theta'\in\Theta'$ such that $\abs{\log P_\theta(y|x)-\log P_{\theta'}(y|x)}\leq \alpha$ for all $x\in\cX, y\in\cY$, and hence $\abs{L\ind{t}(\theta)-L\ind{t}(\theta')}\leq \alpha$ almost surely. Therefore, combining the results above and using $\log w\leq w-1$ for $w>0$, we have
\begin{align*}
    \sum_{t=1}^n \EE\cond{ \DHs{ P_\theta(\cdot|x\ind{t}), P_\ths(\cdot|x\ind{t}) } }{\filt\ind{t-1}}
    \leq&~ \sum_{t=1}^{n} -\log \cond{\exp\paren{ \frac12 L\ind{t}(\theta)}}{\cF\ind{t-1}} \\
    \leq&~ n\alpha + \sum_{t=1}^{n} -\log \cond{\exp\paren{ \frac12 L\ind{t}(\theta')}}{\cF\ind{t-1}} \\
    \leq&~ n\alpha - \frac12 \sum_{t=1}^{n} L\ind{t}(\theta') +\log(N/\delta) \\
    \leq&~ 2n\alpha - \frac12 \sum_{t=1}^{n} L\ind{t}(\theta) +\log(N/\delta).
\end{align*}
By the arbitrariness of $\theta\in\Theta$, the proof is completed.
\end{proofof}

\begin{lemma}[{\citet[Lemma A.4]{foster2021statistical}}]\label{lemma:concen}
For any sequence of real-valued random variables $X\ind{1},\cdots,X\ind{T}$ adapted to a filtration $\left(\filt\ind{t}\right)_{t=1}^T$, it holds that with probability at least $1-\delta$, for all $n\in[T]$,
$$
\sum_{t=1}^{n} -\log \cond{\exp(-X\ind{t})}{\cF\ind{t-1}} \leq \sum_{t=1}^{n} X\ind{t} +\log \left(1/\delta\right).
$$
\end{lemma}

\section{Missing Proofs in \pref{sec: upper-bound-cov}}\label{app: upper-bound-cov}

\subsection{Proof of \cref{thm: model-free-thm}}\label{appdx:proof-cov-upper}

We first present a more detailed statement of the upper bound of \cref{thm: model-free-thm}, as follows.
\begin{theorem}\label{thm: model-free-thm-full}
Let $\delta\in(0,1)$, $\rho\in[0,1)$, and we denote $\ccov=\ccov(\Pi_\cF)$, where $\Pi_\cF=\set{\pi_f:f\in\cF}$ is the policy class induced by $\cF$. Suppose that \cref{asmp:realizable} and \pref{asmp:complete} hold. Then \whp, the output policy $\pihat$ of \cref{alg: model-free} satisfies
\begin{align*}
    \Vs(s_1)-V^{\pihat}(s_1) =&~ \frac1T \sum_{t=1}^T\paren{ \Vs(s_1)-V^{\pi^\iter{t}}(s_1) } \\
    \leq&~ \bigO{H}\cdot \brac{ \frac{\log (N(\rho)/\delta)+TH^2(\rho+\epsapp^2)}{\lambda}+ \frac{\lambda \ccov\log(T)}{T} }.
\end{align*}
Therefore, for any $\eps\in(0,1)$, with the optimally-tuned parameter $\lambda$, it holds that $\Vs(s_1)-V^{\pihat}(s_1)\leq \eps+\tbO{\sqrt{\ccov}H^2\epsapp}$, %
as long as
\begin{align*}
    T\geq \tbO{\frac{\ccov H^2}{\eps^2}\cdot \log N(\eps^2/(\ccov H^4))}.
\end{align*}
\end{theorem}

Recall that we let $\Qsp\in\cQ, \Rsp\in\cR$ be such that
\begin{align*}
    \max_{h\in[H]}\linf{\Qsp_h-\Qs_h}\leq \epsapp, \qquad \max_{h\in[H]}\linf{\Rsp_h-\Rs_h}\leq \epsapp.
\end{align*}
For each $t\in[T]$, we write $\cD\ind{t}$ to be the dataset maintained by \cref{alg: model-free} at the end of the $t$th iteration, i.e.,
\begin{align*}
\cD\ind{t}=\set{ (\tau^\iter{k,h},r^\iter{k,h}) }_{k\leq t, h\in[H]}.    
\end{align*}

We summarize the uniform concentration results for the loss $\LBE[\cD\ind{t}]$ and $\LRM[\cD\ind{t}]$ as follows. We note that these concentration bounds are fairly standard (see e.g.~\citet{jin2021bellman}), and for completeness, we present the proof in \cref{appdx:proof-all-loss-concen}.
\begin{proposition}\label{prop:all-loss-concen}
Let $\delta\in(0,1), \rho\geq 0$. Suppose that \cref{asmp:realizable} and \cref{asmp:complete} holds. 
Then \whp, for all $t\in[T], f\in\cF, R\in\cR$, it holds that
\begin{align*}
    &~\frac{1}{2} \sum_{k\leq t} \sum_{h=1}^H \EE^{\pi^\iter{k}\circ_h \piexp} \paren{ R(\tau) - \Rs(\tau) }^2
    \leq \LRM[\cD^\iter{t}](R)-\LRM[\cD^\iter{t}](\Rsp)+H\kappa, \\
    &~\frac12 \sum_{k\leq t} \sum_{h=1}^{H} \EE^{\pi^\iter{k}} \paren{ f_h(s_h,a_h)-[\BERh f_{h+1}](s_h,a_h)}^2
    \leq \LBE[\cD^\iter{t}](f;R)-\LBE[\cD^\iter{t}](\Qsp;\Rsp) +H\kappa,
\end{align*}
where
\begin{align*}
    \kappa=C\paren{\log N(\rho)+\log(H/\delta)+TH^2(\epsapp^2+\rho)},
\end{align*}
and $C>0$ is an absolute constant.
\end{proposition}

\paragraph{Performance difference decomposition}
Denote $\Vsp(s_1)\defeq \max_{a\in\cA} \Qsp(s_1,a)$. Then it is clear that $\abs{\Vsp(s_1)-\Vs(s_1)}\leq \epsapp$. Therefore, for any $t\in[T]$, by optimism (the definition of $(f^\iter{t},R^\iter{t})$), it holds that
\begin{align*}
    &~\Vs(s_1)-\epsapp
    \leq \Vsp(s_1) \\
    =&~ \Vsp(s_1)-\frac{\LBE[\cD^\iter{t-1}](\Qsp,\Rsp) + \LRM[\cD^\iter{t-1}](\Rsp)}{\lambda} + \frac{\LBE[\cD^\iter{t-1}](\Qsp,\Rsp) + \LRM[\cD^\iter{t-1}](\Rsp)}{\lambda}\\
    \leq&~ f_1^\iter{t}(s_1,\pi^\iter{t})-\frac{\LBE[\cD^\iter{t-1}](f^\iter{t},R^\iter{t}) + \LRM[\cD^\iter{t-1}](R^\iter{t})}{\lambda} + \frac{\LBE[\cD^\iter{t-1}](\Qsp,\Rsp) + \LRM[\cD^\iter{t-1}](\Rsp)}{\lambda}.
\end{align*}
Furthermore, by the standard performance difference lemma~\citep{kakade2002approximately}, it holds that
\begin{align}\label{eq:perf-diff}
    f_1^\iter{t}(s_1,\pi^\iter{t})-V^{\pi^\iter{t}}(s_1)
    =&~ \sum_{h=1}^H \EE^{ \pi^\iter{t} }\brac{ f_h^\iter{t}(s_h,a_h)-[\BE_hf_{h+1}^\iter{t}](s_h,a_h) }.
\end{align}
Based on \eqref{eq:perf-diff}, the existing approaches (with per-step reward feedback) bound the expectation of the Bellman error $e^\iter{t}_h(s_h,a_h)\defeq f_h^\iter{t}(s_h,a_h)-[\BE_hf_{h+1}](s_h,a_h)$ through various arguments (e.g., eluder argument~\citep{jin2021bellman} and coverability argument~\citep{xie2022role}). However, in outcome reward model, it is possible that $\EE^{ \pi^\iter{t} }\brac{ f_h^\iter{t}(s_h,a_h)-[\BE_hf_{h+1}](s_h,a_h) }$ is large even when the sub-optimality of $\pi^\iter{t}$ is small, because the outcome reward is invariant under shifting of the ground-truth reward function $\Rs$. 

Therefore, we consider the following refined decomposition:
\begin{align}\label{eq:perf-diff-traj}
\begin{aligned}
    f_1^\iter{t}(s_1)-V^{\pi^\iter{t}}(s_1)
    =&~ \sum_{h=1}^H \EE^{ \pi^\iter{t} }\brac{ f_h^\iter{t}(s_h,a_h)-[\BER[R^\iter{t}]f_{h+1}^\iter{t}](s_h,a_h) } \\
    &~+ \EE^{ \pi^\iter{t} }\brac{ \sum_{h=1}^H R^\iter{t}_h(s_h,a_h)-\sum_{h=1}^H\Rs_h(s_h,a_h) }.
\end{aligned}
\end{align}

\paragraph{Coverability argument}
Following the coverability argument of \citet{xie2022role}, we have the following upper bound on the expected Bellman errors.
\begin{proposition}\label{prop:cov-Bellman-err}
Denote $e^\iter{t}_h\defeq f_h^\iter{t}-\BE_hf_{h+1}$. Then, for each $h\in[H]$, it holds that
\begin{align*}
    \sum_{t=1}^T \EE^{ \pi^\iter{t} }\abs{e_h^\iter{t}(s_h,a_h)} 
    \leq&~ \sqrt{2\ccov\log\paren{1+\frac{\ccov T}{\kappa}}\cdot \brac{ 2T\kappa+\sum_{1\leq k<t\leq T} \EE^{\pi^\iter{k}} e_h^\iter{t}(s_h,a_h)^2 }}.
\end{align*}
\end{proposition}

Following the ideas of \citet{jia2025we}, we prove the following upper bound on the reward errors.
\begin{proposition}\label{prop:cov-RM-err}
It holds that
\begin{align*}
    &~\sum_{t=1}^T\EE^{ \pi^\iter{t} }\abs{R^\iter{t}(\tau)-\Rs(\tau) } \\
    \leq&~ \sqrt{8H\ccov\log\paren{1+\frac{\ccov T}{\kappa}}\cdot \brac{ HT\kappa+\sum_{1\leq k<t\leq T} \sum_{h=1}^H \EE^{\pi^\iter{k}\circ_h \piexp} \paren{ R^\iter{t}(\tau) - \Rs(\tau) }^2 }}.
\end{align*}
\end{proposition}

Based on the results above, we finalize the proof of \cref{thm: model-free-thm}.
\begin{proofof}{\cref{thm: model-free-thm}}
By optimism and the decomposition \eqref{eq:perf-diff-traj}, it holds that for $t\in[T]$,
\begin{align*}
    \Vs(s_1)-V^{\pi^\iter{t}}(s_1)-\epsapp
    \leq &~ \sum_{h=1}^H \EE^{ \pi^\iter{t} }\brac{ e_h^\iter{t}(s_h,a_h)} - \frac{\LBE[\cD^\iter{t-1}](f^\iter{t},R^\iter{t})-\LBE[\cD^\iter{t-1}](\Qsp,\Rsp)}{\lambda} \\
    &~+ \EE^{ \pi^\iter{t} }\brac{ R^\iter{t}(\tau)-\Rs(\tau) } - \frac{ \LRM[\cD^\iter{t-1}](R^\iter{t})- \LRM[\cD^\iter{t-1}](\Rsp)}{\lambda}.
\end{align*}
Then, under the success event of \cref{prop:all-loss-concen}, we have
\begin{align}\label{eq:proof-subopt-t}
\begin{aligned}
    &~ \Vs(s_1)-V^{\pi^\iter{t}}(s_1)-\epsapp-\frac{2H\kappa}{\lambda} \\
    \leq &~ \sum_{h=1}^H \paren{ \EE^{ \pi^\iter{t} }\brac{ e_h^\iter{t}(s_h,a_h)} - \frac{1}{2\lambda}\sum_{k<t} \EE^{\pi^\iter{k}} e_h^\iter{t}(s_h,a_h)^2 }\\
    &~+ \EE^{ \pi^\iter{t} }\brac{ R^\iter{t}(\tau)-\Rs(\tau) } - \frac{1}{2\lambda}\sum_{k<t} \sum_{h=1}^H \EE^{\pi^\iter{k}\circ_h \piexp} \paren{ R^\iter{t}(\tau) - \Rs(\tau) }^2.
\end{aligned}
\end{align}
Applying \cref{prop:cov-Bellman-err} and Cauchy inequality gives for all $h\in[H]$,
\begin{align*}
    \sum_{t=1}^T \EE^{ \pi^\iter{t} }\abs{e_h^\iter{t}(s_h,a_h)} 
    \leq&~ \lambda \ccov\log\paren{1+\frac{\ccov T}{\kappa}} + \frac{1}{2\lambda }\brac{ 2T\kappa+\sum_{1\leq k<t\leq T} \EE^{\pi^\iter{k}} e_h^\iter{t}(s_h,a_h)^2 },
\end{align*}
and similarly, applying \cref{prop:cov-RM-err} and Cauchy inequality gives
\begin{align*}
    &~\sum_{t=1}^T\EE^{ \pi^\iter{t} }\abs{R^\iter{t}(\tau)-\Rs(\tau) } \\
    \leq&~ 4\lambda H\ccov\log\paren{1+\frac{\ccov T}{\kappa}}+\frac{1}{2\lambda }\brac{ HT\kappa+\sum_{1\leq k<t\leq T} \sum_{h=1}^H \EE^{\pi^\iter{k}\circ_h \piexp} \paren{ R^\iter{t}(\tau) - \Rs(\tau) }^2 }.
\end{align*}
Therefore, we take summation of \eqref{eq:proof-subopt-t} over $t\in[T]$, and combining the inequalities above gives
\begin{align*}
    \sum_{t=1}^T\paren{ \Vs(s_1)-V^{\pi^\iter{t}}(s_1) } \leq T\epsapp+\frac{4TH\kappa}{\lambda}+5\lambda H\ccov\log\paren{1+\frac{\ccov T}{\kappa}}.
\end{align*}
This is the desired upper bound.
\end{proofof}

\subsection{Proof of \cref{prop:cov-Bellman-err} and \cref{prop:cov-RM-err}}

The following proposition is an generalized version of the results in \citet[Appendix D]{xie2022role}. For proof, see e.g. \citet{chen2024near}.
\begin{proposition}[{\cite{xie2022role}}]\label{prop:coverage-eluder}
Let $C\geq 1$ be a parameter. Suppose that $p\ind{1},\cdots,p\ind{T}$ is a sequence of distributions over $\cX$, and there exists $\mu\in\Delta(\cX)$ such that $p\ind{t}(x)/\mu(x)\leq C$ for all $x\in\cX$, $t\in[T]$. Then for any sequence $\psi\ind{1},\cdots,\psi\ind{T}$ of functions $\cX\to[0,1]$ and constant $B\geq 1$, it holds that
\begin{align*}
    \sum_{t=1}^T \EE_{x\sim p\ind{t}} \psi\ind{t}(x) \leq \sqrt{2C\log\paren{1+\frac{C T}{B}}\brac{2TB+\sum_{t=1}^T \sum_{k<t} \EE_{x\sim p\ind{k}} \psi\ind{t}(x)^2} }.
\end{align*}
\end{proposition}
As a warm-up, we prove \cref{prop:cov-Bellman-err} by directly invoking \cref{prop:coverage-eluder}.

\begin{proofof}{\cref{prop:cov-Bellman-err}}
Fix a $h\in[H]$. To apply \cref{prop:coverage-eluder}, we consider $\cX=\cS\times\cA$, and define
\begin{align*}
    p\ind{t}\defeq \PP^{\pi^\iter{t}}\paren{(s_h,a_h)=\cdot}\in\Delta(\cS\times\cA), \qquad 
    \psi\ind{t}\defeq \abs{e^\iter{t}_h}\in(\cS\times\cA\to[0,1]).
\end{align*}
By the definition of coverability (\cref{def: coverability}), for $C=\ccov(\Pi)$, there exists $\mu\in\Delta(\cS\times\cA)$ such that $p\ind{t}(s,a)/\mu(s,a)\leq C$ for all $(s,a)\in\cS\times\cA$. 
Therefore, applying \cref{prop:coverage-eluder} with $B=\kappa\geq 1$ gives the desired upper bound.
\end{proofof}

Next, we proceed to prove \cref{prop:cov-RM-err}. Our key proof technique is summarized in the following proposition, which is inspired by the (rather sophisticated) analysis of \citet{jia2025we}.

\newcommand{\Dbar}{\overline{D}}
\begin{proposition}\label{lem:traj-decomp}
Recall that for any $D=(D_h:\cS\times\cA\to\R)$, we denote
\begin{align*}
    D(\tau)=\sum_{h=1}^H D_h(s_h,a_h), \qquad \forall \tau=(s_1,a_1,\cdots,s_H,a_H)\in(\cS\times\cA)^H.
\end{align*}
Fix a Markov policy $\piexp$, we denote $\Dbar_1(s)=\Dbar_{H+1}(s)=0$, and
\begin{align*}
    \Dbar_h(s)\defeq \EE^{\piexp}\cond{\sum_{\ell=h}^H D_\ell(s_\ell,a_\ell)}{s_h=s}, \qquad \forall 1<h\leq H, s\in\cS.
\end{align*}
Then for any policy $\pi$, it holds that
\begin{align*}
    \sum_{h=1}^H \EE^\pi\paren{ D_h(s_h,a_h)+\Dbar_{h+1}(s_{h+1})-\Dbar_h(s_h) }^2\leq 4\sum_{h=1}^H \EE^{\pi\circ_h \piexp} D(\tau)^2.
\end{align*}
\end{proposition}

The proof of \cref{lem:traj-decomp} is deferred to the end of this subsection. With \cref{lem:traj-decomp}, we prove \cref{prop:cov-RM-err} as follows.
\begin{proofof}{\cref{prop:cov-RM-err}}
To apply \cref{lem:traj-decomp}, for each $t\in[T]$, we consider
\begin{align*}
    \Delta_h\ind{t}(s)\defeq \EE^{\piexp}\cond{\sum_{\ell=h}^H R_\ell^{\iter{t}}(s_\ell,a_\ell)-\sum_{\ell=h}^H \Rs_\ell(s_\ell,a_\ell)}{s_h=s}, \quad\forall h=2,\cdots,H, s\in\cS,
\end{align*}
Then, by \cref{lem:traj-decomp}, it holds that for any policy $\pi$,
\begin{align}\label{eq:proof-cov-RM-1}
\begin{aligned}
    &\sum_{h=1}^H \EE^\pi\paren{ R_h^\iter{t}(s_h,a_h)-\Rs_h(s_h,a_h)+\Delta_{h+1}\ind{t}(s_{h+1})-\Delta_h\ind{t}(s_h) }^2 \\
    &\qquad \leq 4 \sum_{h=1}^H \EE^{\pi\circ_h \piexp} \paren{ R^\iter{t}(\tau) - \Rs(\tau) }^2.
\end{aligned}
\end{align}
Furthermore,
\begin{align}\label{eq:proof-cov-RM-2}
\begin{aligned}
     \EE^{ \pi }\abs{R^\iter{t}(\tau)-\Rs(\tau) }
 =&~ \EE^{ \pi }\abs{\sum_{h=1}^H \brac{ R_h^\iter{t}(s_h,a_h)-\Rs_h(s_h,a_h)+\Delta_{h+1}\ind{t}(s_{h+1})-\Delta_h\ind{t}(s_h) } } \\
\leq&~
    \sum_{h=1}^H \EE^\pi\abs{ R_h^\iter{t}(s_h,a_h)-\Rs_h(s_h,a_h)+\Delta_{h+1}\ind{t}(s_{h+1})-\Delta_h\ind{t}(s_h) }.
\end{aligned}
\end{align}
Therefore, to apply \cref{prop:coverage-eluder}, we consider the space $\cX=\cS\times\cA\times\cS$, and for each $h\in[H]$, we define
\begin{align*}
    p\ind{t}_h\defeq&~ \PP^{\pi^\iter{t}}\paren{(s_h,a_h,s_{h+1})=\cdot}\in\Delta(\cS\times\cA\times\cS), \\ 
    \psi\ind{t}_h(s,a,s')\defeq&~ \abs{ R_h^\iter{t}(s,a)-\Rs_h(s,a)+\Delta_{h+1}\ind{t}(s')-\Delta_h\ind{t}(s) }.
\end{align*}
Note that for any $h$, we have $\psi\ind{t}_h:\cS\times\cA\times\cS\to [0,1]$. Further, let $\mu_h\in\Delta(\cS\times\cA)$ be the distribution such that $\linf{d^\pi_h/\mu_h}\leq \ccov$ for any policy $\pi$. Then we can consider the distribution $\omu_h\in\Delta(\cS\times\cA\times\cS)$ given by $\omu_h(s,a,s')=\mu_h(s,a)\TT_h(s'|s,a)$. Then it holds that
\begin{align*}
    \linf{\frac{p\ind{t}_h}{\omu_h}}=\sup_{s,a,s'} \frac{p\ind{t}_h(s,a,s')}{\omu_h(s,a,s')}=\sup_{s,a,s'} \frac{d^{\pi^\iter{t}}_h(s,a)\TT_h(s'|s,a)}{\mu_h(s,a)\TT_h(s'|s,a)}\leq \ccov.
\end{align*}
Therefore, for $h\in[H]$, applying \cref{prop:coverage-eluder} on the sequence $(p\ind{1}_h,\cdots,p\ind{T}_h)$ and $(\psi\ind{1}_h,\cdots,\psi\ind{T}_h)$ gives
\begin{align*}
    \sum_{t=1}^T \EE_{x\sim p_h\ind{t}} \psi_h\ind{t}(x) \leq \sqrt{2\ccov \log\paren{1+\frac{\ccov T}{\kappa}}\brac{2T\kappa+\sum_{t=1}^T \sum_{k<t} \EE_{x\sim p_h\ind{k}} \psi_h\ind{t}(x)^2} }.
\end{align*}
To conclude, we combine the inequalities above and bound
\begin{align*}
    &~\sum_{t=1}^T\EE^{ \pi^\iter{t} }\abs{R^\iter{t}(\tau)-\Rs(\tau) } \\
    \leq&~ \sum_{t=1}^T\sum_{h=1}^H \EE^{\pi^\iter{t}} \abs{ R_h^\iter{t}(s_h,a_h)-\Rs_h(s_h,a_h)+\Delta_{h+1}\ind{t}(s_{h+1})-\Delta_h\ind{t}(s_h) } \\
    =&~ \sum_{t=1}^T\sum_{h=1}^H \EE_{x\sim p_h\ind{t}} \psi_h\ind{t}(x) 
    =\sum_{h=1}^H \sum_{t=1}^T \EE_{x\sim p_h\ind{t}} \psi_h\ind{t}(x) \\
    \leq&~ \sum_{h=1}^H \sqrt{2\ccov \log\paren{1+\frac{\ccov T}{\kappa}}\brac{2T\kappa+\sum_{t=1}^T \sum_{k<t} \EE_{x\sim p_h\ind{k}} \psi_h\ind{t}(x)^2} } \\
    \leq&~ \sqrt{2H\ccov \log\paren{1+\frac{\ccov T}{\kappa}}\brac{2TH\kappa+\sum_{h=1}^H \sum_{t=1}^T \sum_{k<t} \EE_{x\sim p_h\ind{k}} \psi_h\ind{t}(x)^2} } \\
    \leq&~ \sqrt{8H\ccov\log\paren{1+\frac{\ccov T}{\kappa}}\cdot \brac{ HT\kappa+\sum_{1\leq k<t\leq T} \sum_{h=1}^H \EE^{\pi^\iter{k}\circ_h \piexp} \paren{ R^\iter{t}(\tau) - \Rs(\tau) }^2 }},
\end{align*}
where the first inequality follows from \eqref{eq:proof-cov-RM-2}, the second last line follows from Cauchy inequality, and last inequality follows from the definition of $(p_h\ind{t},\psi_h\ind{t})$ and \eqref{eq:proof-cov-RM-1}.
\end{proofof}

\begin{proofof}{\cref{lem:traj-decomp}}
For $\tau=(s_1,a_1,\cdots,s_H,a_H)\in(\cS\times\cA)^H$, we denote $\tau_h=(s_1,a_1,\cdots,s_{h},a_h,s_{h+1})$ to be the prefix sequence of $\tau$ for each $h\in[H]$. Then, we note that
\begin{align*}
    \EE^{\pi\circ_h\piexp}\cond{ D(\tau) }{\tau_h}
    =&~\sum_{\ell=1}^h D_\ell(s_\ell,a_\ell)+\EE^{\pi\circ_h\piexp}\cond{ \sum_{\ell=h+1}^H D_\ell(s_\ell,a_\ell) }{\tau_h} \\
    =&~\sum_{\ell=1}^h D_\ell(s_\ell,a_\ell)+\Dbar_{h+1}(s_{h+1}),
\end{align*}
because the policy $\pi\circ_h\piexp$ executes the Markov policy $\piexp$ starting at the $(h+1)$-th step. Therefore, it holds that
\begin{align*}
    \EE_{\tau_h\sim \pi}\paren{\sum_{\ell=1}^h D_\ell(s_\ell,a_\ell)+\Dbar_{h+1}(s_{h+1})}^2
    =&~
    \EE_{\tau_h\sim \pi\circ_h \piexp} \paren{\EE^{\pi\circ_h\piexp}\cond{ D(\tau) }{\tau_h}}^2 \\
    \leq&~ \EE_{\tau\sim \pi\circ_h \piexp} D(\tau)^2
    =\EE^{\pi\circ_h \piexp} D(\tau)^2,
\end{align*}
where the first equality follows from the fact that the policy $\pi\circ_h\piexp$ executes $\pi$ for the first $h$ steps. Therefore, for $h>1$, it holds that
\begin{align*}
    &~\EE^\pi\paren{ D_h(s_h,a_h)+\Dbar_{h+1}(s_{h+1})-\Dbar_h(s_h) }^2 \\
    =&~ \EE_{\tau_h\sim \pi}\paren{\sum_{\ell=1}^h D_\ell(s_\ell,a_\ell)+\Dbar_{h+1}(s_{h+1})-\sum_{\ell=1}^{h-1} D_\ell(s_\ell,a_\ell)-\Dbar_{h}(s_{h})}^2 \\
    \leq&~ 2\EE_{\tau_h\sim \pi}\paren{\sum_{\ell=1}^h D_\ell(s_\ell,a_\ell)+\Dbar_{h+1}(s_{h+1})}^2+\EE_{\tau_{h-1}\sim \pi}\paren{\sum_{\ell=1}^{h-1} D_\ell(s_\ell,a_\ell)+\Dbar_{h}(s_{h})}^2 \\
    \leq&~ 2\EE^{\pi\circ_h \piexp} D(\tau)^2+\EE^{\pi\circ_{h-1} \piexp} D(\tau)^2.
\end{align*}
For $h=1$, because $\Dbar_1(s)=0$, we already have
\begin{align*}
    \EE^\pi\paren{ D_1(s_1,a_1)+\Dbar_{2}(s_{2})-\Dbar_1(s_1) }^2
    =\EE^\pi\paren{ D_1(s_1,a_1)+\Dbar_{2}(s_{2}) }^2 \leq \EE^{\pi\circ_1 \piexp} D(\tau)^2.
\end{align*}
Taking summation over $h\in[H]$ completes the proof.
\end{proofof}

\subsection{Proof of \cref{prop:all-loss-concen}}\label{appdx:proof-all-loss-concen}

We prove \cref{prop:all-loss-concen} in \cref{lem:RM-loss-concen} and \cref{lem:BE-loss-concen} separately. Recall again that under \cref{asmp:realizable}, the function $\Qsp\in\cF, \Rsp\in\cR$ satisfy
\begin{align*}
    \max_{h\in[H]}\linf{\Qsp_h-\Qs_h}\leq \epsapp, \qquad \max_{h\in[H]}\linf{\Rsp_h-\Rs_h}\leq \epsapp.
\end{align*}

\begin{lemma}\label{lem:RM-loss-concen}
Under \cref{asmp:realizable}, \whp, for any $t\in[T]$, for all $R\in\cR$, it holds that
\begin{align*}
    \frac{1}{2} \sum_{k\leq t} \sum_{h=1}^H \EE^{\pi^\iter{k}\circ_h \piexp} \paren{ R(\tau) - \Rs(\tau) }^2
    \leq&~ \LRM[\cD^\iter{t}](R)-\LRM[\cD^\iter{t}](\Rsp) \\
    &~+ 15H\log N(\rho)+15(2/\delta)+2TH^3\epsapp^2+4TH^2\rho.
\end{align*}
\end{lemma}

\begin{proofof}{\cref{lem:RM-loss-concen}}
To apply \cref{lem:ERM-unif}, we consider the whole history
\begin{align*}
    \set{(\tau^\iter{t,h},r^\iter{t,h})}_{t\in[T],h\in[H]}.
\end{align*}
generated by executing \cref{alg:golf}, and recall that $\cD\ind{t-1}=\set{(\tau^\iter{k,h},r^\iter{k,h})}_{k<t,h\in[H]}$ 
is the history up to the $t$-th iteration.
Note that $(\tau^\iter{t,1},r^\iter{t,1}),\cdots,(\tau^\iter{t,H},r^\iter{t,H})$ are pairwise independent given $\cD\ind{t-1}$, with
\begin{align*}
    \tau^\iter{t,h}\sim \pi^\iter{t}\circ_h \piexp, \qquad \EE\cond{r^\iter{t,h}}{\cD\ind{t-1},\tau^\iter{t,h}}=\Rs(\tau^\iter{t,h}).
\end{align*}
Also note that $r\in[0,1]$ almost surely, and we regard $\cR\subseteq ((\cS\times\cA)^H\to[0,1])$, and $\Rsp\in\cR$ satisfies $\abs{\Rsp(\tau)-\Rs(\tau)}\leq H\epsapp$ for all $\tau\in(\cS\times\cA)^H$.

Therefore, applying \cref{lem:ERM-unif} on the function class $\cR$ and the sequence $\set{(\tau^\iter{t,h},r^\iter{t,h})}_{t\in[T],h\in[0,H]}$ gives that \whp, for all $R\in\cR$, $t\in[T]$, it holds that %
\begin{align*}
    &~ \frac{1}{2} \sum_{k=1}^t \sum_{h=1}^H \EE^{\pi^\iter{k}\circ_h \piexp} \paren{ R(\tau) - \Rs(\tau) }^2 
    = \frac{1}{2} \sum_{k=1}^t \sum_{h=1}^H \EE\cond{ \paren{ R(\tau^\iter{k,h}) - \Rs(\tau^\iter{k,h}) }^2 }{ \cD\ind{k-1} } \\
    \leq&~ \LRM[\cD^\iter{t}](R)-\LRM[\cD^\iter{t}](\Rsp) + 15\log(2N(\cR,H\rho)/\delta)+2TH^3\epsapp^2+4TH^2\rho.
\end{align*}
Finally, we note that $\log N(\cR,H\rho)\leq \sum_{h=1}^H \log N(\cR_h,\rho)\leq H\log N(\rho)$. This gives the desired upper bound.
\end{proofof}

Similarly, we prove \cref{prop:all-loss-concen} (2) as follows, following \citet{jin2021bellman}.

\begin{lemma}\label{lem:BE-loss-concen}
Fix $h\in[H]$ and $\delta\in(0,1)$, $\rho\geq 0$.
Suppose that \cref{asmp:realizable} and \cref{asmp:complete} holds. Then \whp, the following holds:

(1) For each $t\in[T]$,
\begin{align*}
    \LBEt(\Qsp_h,\Qsp_{h+1};\Rsp) - \inf_{g_h\in\cG_h} \LBEt(g_h, \Qsp_{h+1}; \Rsp)\leq \bigO{TH\epsapp^2+TH\rho+\log(N(\rho)/\delta)}.
\end{align*}

(2) For each $t\in[T]$, for all $f_h\in\cF_h, f_{h+1}\in\cF_{h+1}$, and $R_h\in\cR_h$, 
\begin{align*}
    &~ \frac12 \sum_{k=1}^t \EE^{\pi^\iter{k}} \paren{ f_h(s_h,a_h)-[\BERh f_{h+1}](s_h,a_h)}^2 \\
    \leq&~ \LBEt(f_h,f_{h+1};R_h) - \inf_{g_h\in\cG_h} \LBEt(g_h, f_{h+1}; R_h) 
    +\bigO{TH\epsapp^2+TH\rho+\log(N(\rho)/\delta)},
\end{align*}
where we use $\bigO{\cdot}$ to hide absolute constant for simplicity. \cfcomment{updated}
\end{lemma}

\begin{proofof}{\cref{lem:BE-loss-concen}}
Fix $h\in[H]$ and denote $N\defeq N(\rho)$. We let $\cF'_{h+1}$ be a minimal $\rho$-covering of $\cF_{h+1}$, and let $\cR'_h$ be a minimal $\rho$-covering of $\cR_h$. By definition, $|\cF_{h+1}'|\leq N, |\cR_h'|\leq N$.

In the following, we adopt the notation of the proof of \cref{lem:RM-loss-concen}. Recall that conditional on $\cD\ind{t-1}$, 
\begin{align*}
    \tau^\iter{t,\ell}=(s_1^\iter{t,\ell},a_1^\iter{t,\ell},\cdots,s_H^\iter{t,\ell},a_H^\iter{t,\ell}) \sim \pi^\iter{t}\circ_\ell \piexp,
\end{align*}
and $\tau^\iter{t,1},\cdots,\tau^\iter{t,H}$ are independent conditional on $\cD\ind{t-1}$.
For simplicity, we denote $x\ind{t,\ell}\defeq (s_h^\iter{t,\ell}, a_h^\iter{t,\ell})$. 

Fix $f_{h+1}\in\cF_{h+1}'\cup\set{\Qsp_{h+1}}$ and $R_h\in\cR_h'\cup\set{\Rsp_{h}}$, we consider %
\begin{align*}
    y\ind{t,\ell}\defeq f_{h+1}(s_{h+1}^\iter{t,\ell})+R_h(s_h^\iter{t,\ell}, a_h^\iter{t,\ell}).
\end{align*}
and it holds that
\begin{align*}
    \EE\cond{y\ind{t,\ell}}{\cD\ind{t-1},x\ind{t,\ell}}=[\BERh f_{h+1}](x\ind{t,\ell}).
\end{align*} 
Then, for any $g_h\in\cG_h$, it holds that
\begin{align*}
    \LBEt(g_h, f_{h+1}; R_h)=\sum_{k=1}^t \sum_{\ell=1}^H \paren{g_h(x\ind{t,\ell})-y\ind{t,\ell}}^2,
\end{align*}
and we also have
\begin{align*}
    \sum_{k=1}^t\sum_{\ell=1}^H \EE^{\pi^\iter{k}\circ_\ell \piexp}\paren{ g_h(s_h,a_h)-[\BERh f_{h+1}](s_h,a_h)}^2
    = 
    \sum_{k=1}^t\sum_{\ell=1}^H \EE\cond{\paren{ g_h(x\ind{k,\ell})-[\BERh f_{h+1}](x\ind{k,\ell})}^2}{\cD\ind{k-1}}
\end{align*}
Then, applying \cref{lem:ERM-unif} with the function class $\cH=\cG_h$ yields that with probability at least $1-\frac{\delta}{2N}$, the following holds:

(a) For each $t\in [T]$, for any $g_h\in\cG_h$,
\begin{align*}
    &\frac{1}{2}\sum_{k=1}^t \EE^{\pi^\iter{k}}\paren{ g_h(s_h,a_h)-[\BERh f_{h+1}](s_h,a_h)}^2 \\
    \leq& \LBEt(g_h, f_{h+1}; R_h)-\inf_{g_h'\in\cG_h}\LBEt(g_h', f_{h+1}; R_h)+\bigO{\log(N/\delta)+TH\epsapp^2+TH\rho}.
\end{align*}

(b) When $f_{h+1}=\Qsp_{h+1}$ and $R_h=\Rsp_h$, the function $\Qsp_h\in\cF_h\subseteq \cG_h$ satisfies the inequality $\linfn{\Qsp_h-\BERh[\Rsp] \Qsp_{h+1}}\leq 3\epsapp$, and thus
\begin{align*}
    \LBEt(\Qsp_h, \Qsp_{h+1}; \Rsp_h)\leq \inf_{g_h\in\cG_h}\LBEt(g_h, \Qsp_{h+1}; \Rsp_h)+\bigO{\log(N/\delta)+TH\epsapp^2+TH\rho}.
\end{align*}

Therefore, taking the union bound, we know that the inequalities (a) and (b) above hold simultaneously \whp~for all $f_{h+1}\in\cF_{h+1}'\cup\set{\Qsp_{h+1}}$ and $R_h\in\cR_h'\cup\set{\Rsp_{h}}$. In particular, we have completed the proof of (1).

To prove (2), we only need to note that $\cG_h\subseteq \cF_h$, and for any $f_{h+1}\in\cF_{h+1}, R_h\in\cR_{h}$, there exists $f_{h+1}'\in\cF_{h+1}', R_h'\in\cR_h'$ such that $\linfn{f_{h+1}-f_{h+1}'}\leq \rho$, $\linfn{R_h-R_h'}\leq \rho$. Therefore, by the standard covering argument and the fact that $\pi^\iter{k}\circ_H \piexp=\pi^\iter{k}$, we have also shown (2).
\end{proofof}

\section{Proof of \cref{thm: deterministic}}\label{app: upper-bound-DMDP}

In this section, we provide the proof of \cref{thm: deterministic}, which is a direct adaption of the proof of \cref{thm: model-free-thm} in \cref{app: upper-bound-cov}. We first present a more detailed statement of the upper bound (with any parameter $\lambda>0$).
\begin{theorem}\label{thm: deterministic-full}
Suppose that \cref{asmp:realizable} holds. Then \whp, \pref{alg: deterministic} achieves 
\begin{align*}
    \frac1T \sum_{t=1}^T\paren{ \Vs(s_1^\iter{t})-V^{\pi^\iter{t}}(s_1^\iter{t}) } 
    \leq&~ \epsapp + \bigO{1}\cdot \brac{ \frac{H^3\log(\NT{\cF,T}/\delta)+T\epsapp^2}{\lambda}+ \frac{\lambda H\ccov'(\Pi)}{T} },
\end{align*}
\end{theorem}
We also work with a slightly relaxed version of \cref{asmp:deterministic}.
\begin{assumption}\label{asmp:DMDP-relaxed}
Under any policy $\pi$, for each $h\in[H]$, it holds that almost surely
\begin{align*}
    \Qs_h(s_h,a_h)=\Rs_h(s_h,a_h)+\Vs_{h+1}(s_{h+1}).
\end{align*}
\end{assumption}
Further, to simply the notation, for each $f\in\cF$, we recall that the induced reward model $R^f$ is defined as $R^f_1(s,a)\defeq f_1(s,a)$, $R^f_h(s,a)=f_h(s,a)-f_{h}(s)$, which implies
\begin{align*}
    R^f(\tau)&~=\sum_{h=1}^H f_h(s_h,a_h)-f_{h+1}(s_{h+1}).
\end{align*}

\paragraph{Uniform convergence}
For each $t\in[T]$, we define $\cD\ind{t-1}\defeq \set{(\tau^\iter{k},r^\iter{k})}_{k<t}$ be the data collected before $t$th iteration. We also recall that by definition~\eqref{eq:loss-BR}, we have
\begin{align*}
    \LDBE[\cD^\iter{t}](f)\defeq \sum_{k=1}^{t}  \paren{ \sum_{h=1}^H \brac{ f_h(s_h^\iter{k},a_h^\iter{k}) - f_{h+1}(s_{h+1}^\iter{k}) } - r^\iter{k} }^2
    = \sum_{k=1}^{t}  \paren{ R^f(\tau^\iter{k}) - r^\iter{k} }^2.
\end{align*}
Therefore, a direct instantiation of \cref{lem:ERM-unif} on the class $\cR\defeq\set{ R^f: f\in\cF}$ yields the following proposition.
\begin{proposition}\label{prop:DMDP-all-loss-concen}
Let $\delta\in(0,1), \rho\geq 0$. Suppose that \cref{asmp:realizable} and \cref{asmp:DMDP-relaxed} holds. Then \whp, for all $t\in[T], f\in\cF$, it holds that
\begin{align*}
    &~\frac{1}{2} \sum_{k=1}^t \EE^{\pi^\iter{k}} \paren{ R^f(\tau) - \Rs(\tau) }^2
    \leq \LDBE[\cD^\iter{t}](f)-\LBE[\cD^\iter{t}](\Qsp)+\kappa,
\end{align*}
where
\begin{align*}
    \kappa=C\paren{H^3\log(N_{\cF}(\alpha)/\delta)+TH\alpha+T\epsapp^2)},
\end{align*}
$C>0$ is an absolute constant, and we denote $N_{\cF}(\alpha)\defeq \max_{h\in[H]}N(\cF_h,\alpha)$ for any $\alpha\geq 0$.
\end{proposition}

\paragraph{Performance difference decomposition}
In this setting, we can rewrite the decomposition~\eqref{eq:perf-diff} as
\begin{align}\label{eq:perf-diff-DMDP}
\begin{aligned}
f_1^\iter{t}(s_1)-V^{\pi^\iter{t}}(s_1)
    =&~ \sum_{h=1}^H \EE^{ \pi^\iter{t} }\brac{ f_h^\iter{t}(s_h,a_h)-\Rs_h(s_h,a_h)-f_{h+1}^\iter{t}(s_{h+1}) } \\
    =&~ \EE^{ \pi^\iter{t} }\brac{ \sum_{h=1}^H\brac{ f_h^\iter{t}(s_h,a_h)-f_{h+1}^\iter{t}(s_{h+1}) } -\sum_{h=1}^H \Rs_h(s_h,a_h) } \\
    =&~ \EE^{ \pi^\iter{t} }\brac{ R^\iter{t}(\tau)-\Rs(\tau) },
\end{aligned}
\end{align}
where we denote $R^\iter{t}\defeq R^{f^\iter{t}}$, which is a reward model given by 
\begin{align*}
R^\iter{t}_1(s,a)\defeq f_1^\iter{t}(s,a), \qquad R^\iter{t}_h(s,a)=f_h^\iter{t}(s,a)-f_{h}^\iter{t}(s).
\end{align*}

\paragraph{Optimism} Similar to \cref{appdx:proof-cov-upper}, we use the fact that from \eqref{eq:max-f-DMDP},
\begin{align*}
    f^\iter{t}=\max_{f \in \cF} \lambda f_1(s_1^\iter{t}) - \LDBE[\cD^\iter{t-1}](f),
\end{align*}
and hence
\begin{align*}
    \lambda f_1^\iter{t}(s_1^\iter{t}) - \LDBE[\cD^\iter{t-1}](f^\iter{t})\geq \lambda \Vsp_1(s_1^\iter{t})-\LDBE[\cD^\iter{t-1}](f^\iter{t}).
\end{align*}
Using $\abs{\Vsp_1(s_1^\iter{t})-\Vs_1(s_1^\iter{t})}\leq \epsapp$, \eqref{eq:perf-diff-DMDP} and \cref{prop:DMDP-all-loss-concen}, we now deduce that
\begin{align}\label{eq:proof-DMDP-optimism}
\begin{aligned}
    \Vs(s_1^\iter{t})-V^{\pi^\iter{t}}(s_1^\iter{t})\leq&~ \epsapp + \EE^{ \pi^\iter{t} }\brac{ R^\iter{t}(\tau)-\Rs(\tau) } - \frac{ \LDBE[\cD^\iter{t-1}](f^\iter{t})-\LBE[\cD^\iter{t-1}](\Qsp) }{\lambda } \\
    \leq&~ \epsapp + \frac{\kappa}{\lambda} + \EE^{ \pi^\iter{t} }\brac{ R^\iter{t}(\tau)-\Rs(\tau) } - \frac{1}{2\lambda} \sum_{k=1}^{t-1} \EE^{\pi^\iter{k}} \paren{ R^\iter{t}(\tau) - \Rs(\tau) }^2.
\end{aligned}
\end{align}
Therefore, it remains to prove an analogue to \cref{prop:cov-RM-err}.

\paragraph{Coverability argument} We strength \cref{prop:cov-RM-err} using the deterministic nature of the underlying MDP. For each $s\in\cS$ and $h\in[H]$, we define
\begin{align*}
    \cS_h(s;\Pi)\defeq \set{(s',a): \exists \pi\in\Pi,\text{ under $\pi$ and }s_1=s, \text{ it holds that } s_h=s', a_h=a},
\end{align*}
and $N_h(s;\Pi)\defeq |\cS_h(s;\Pi)|$.

\newcommandx{\NCov}[1][1={s_1}]{N(#1)}

\begin{proposition}\label{prop:DMDP-cov-RM-err}
Let $B\geq 1$.
For any initial state $s_1\in\cS$, any sequence of reward functions $R^\iter{1},\cdots,R^\iter{T}$ and any sequence of policies $\pi^\iter{1},\cdots,\pi^\iter{T}$, it holds that
\begin{align*}
    &~\sum_{t=1}^T\EE^{ \pi^\iter{t} }\cond{R^\iter{t}(\tau)-\Rs(\tau)}{s_1} \\
    \leq&~ \sqrt{2\NCov\log\paren{1+\frac{4TH}{B}}\cdot \brac{ 2TB+\sum_{1\leq k<t\leq T} \EE^{\pi^\iter{k}} \cond{ \paren{ R^\iter{t}(\tau) - \Rs(\tau) }^2 } {s_1}}},
\end{align*}
where $N(s_1)\coloneqq \sum_{h=1}^H N_h(s_1; \Pi)$, and the conditional distribution $\EE^{ \pi^\iter{t} }\cond{\cdot}{s_1}$ is taken over the expectation of $\tau$ generated by executing policy $\pi$ starting with the initial state $s_1$.
\end{proposition}
The proof of \cref{prop:DMDP-cov-RM-err} is deferred to the end of this section.

\paragraph{Finalizing the proof}
With the above preparation, we now finalize the proof of \cref{thm: deterministic}.
Taking summation of \eqref{eq:proof-DMDP-optimism} over $t=1,2,\cdots,T$, we have
\begin{align*}
    &~\sum_{t=1}^T \Vs(s_1^\iter{t})-V^{\pi^\iter{t}}(s_1^\iter{t}) \\
    \leq&~ T\epsapp + \frac{T\kappa}{\lambda} + \sum_{t=1}^T\EE^{ \pi^\iter{t} }\brac{ R^\iter{t}(\tau)-\Rs(\tau) } - \frac{1}{2\lambda} \sum_{1\leq k<t\leq T} \EE^{\pi^\iter{k}} \paren{ R^\iter{t}(\tau) - \Rs(\tau) }^2 \\
    =&~ T\epsapp + \frac{T\kappa}{\lambda} + \EE_{s_1\sim \rho}\brac{ \sum_{t=1}^T\EE^{ \pi^\iter{t} }\cond{ R^\iter{t}(\tau)-\Rs(\tau) }{s_1} - \frac{1}{2\lambda} \sum_{1\leq k<t\leq T} \EE^{\pi^\iter{k}} \cond{ \paren{ R^\iter{t}(\tau) - \Rs(\tau) }^2}{s_1}  } \\
    \leq&~ T\epsapp + \frac{2T\kappa}{\lambda} + \EE_{s_1\sim \rho }\brac{ N(s_1)\lambda\log\paren{1+\frac{TH}{\kappa}} },
\end{align*}
where the last inequality follows from \cref{prop:DMDP-cov-RM-err} and Cauchy inequality. This is the desired upper bound.
\qed

\begin{proofof}{\cref{prop:DMDP-cov-RM-err}}
In the following proof, we assume $s_1\in\cS$ is fixed. Consider
\begin{align*}
    \cI\defeq \set{(h,s,a): h\in[H], (s,a)\in\cS_h(s_1;\Pi)}\subseteq [H]\times\cS\times\cA.
\end{align*}
Note that $|\cI|=\sum_{h=1}^H N_h(s_1;\Pi)=\NCov$.  By definition, for any policy $\pi$, there is a unique pair $(s_h^\pi,a_h^\pi)\in \cS_h(s_1;\Pi)$, such that under $\pi$ and starting from $s_1$, we have $s_h=s_h^\pi, a_h=a_h^\pi$ deterministically.

For each $t\in[T]$, we consider the following vectors indexed by $\cI$:
\begin{align*}
    \psi\ind{t}\defeq&~ \brac{ R_h^\iter{t}(s,a)-\Rs_h(s,a) }_{(h,s,a)\in\cI}\in \R^{\cI}, \\
    \phi\ind{t}\defeq&~ \brac{ \PP^{\pi^\iter{t}}(s_h=s,a_h=a|s_1) }_{(h,s,a)\in\cI} = \sum_{h=1}^H e_{(h,s_h^{\pi^\iter{t}},a_h^{\pi^\iter{t}})}\in \R^{\cI}.
\end{align*}
With this definition, it holds that for any $k, t\in[T]$,
\begin{align*}
    \EE^{\pi^\iter{k}} \cond{ R^\iter{t}(\tau) - \Rs(\tau)  } {s_1} = \sum_{h=1}^H \brac{ R^\iter{t}(s_h^{\pi^\iter{k}},a_h^{\pi^\iter{k}})-\Rs(s_h^{\pi^\iter{k}},a_h^{\pi^\iter{k}}) } = \lr \phi\ind{k}, \psi\ind{t} \rr.
\end{align*}
Therefore, we apply the elliptical potential argument~\citep{lattimore2020bandit}. Let $V_t\defeq \sum_{k<t}\phi\ind{k}\paren{\phi\ind{k}}^\top+B\id$. Then it holds that
\begin{align*}
    \sum_{t=1}^T \abs{ \lr \phi\ind{t}, \psi\ind{t} \rr }
    \leq&~ \sum_{t=1}^T \min\set{ \nrm{ \phi\ind{t} }_{V_t^{-1}}, 1} \cdot \max\set{ \nrm{ \psi\ind{t} }_{V_t}, 1} \\
    \leq&~ \sqrt{ \sum_{t=1}^T \min\set{ \nrm{ \phi\ind{t} }_{V_t^{-1}}^2, 1} }\cdot \sqrt{\sum_{t=1}^T \max\set{\nrm{ \psi\ind{t} }_{V_t}^2,1} }.
\end{align*}
Note that
\begin{align*}
    \sum_{t=1}^T \max\set{\nrm{ \psi\ind{t} }_{V_t}^2,1} \leq &~ \sum_{t=1}^T \brac{ 1+B\nrm{\psi\ind{t}}^2+ \sum_{k=1}^{t-1} \lr \phi\ind{k}, \psi\ind{t} \rr^2 } \\
    \leq&~ T(1+4B|\cI|)+\sum_{1\leq k<t\leq T} \EE^{\pi^\iter{k}} \cond{ \paren{ R^\iter{t}(\tau) - \Rs(\tau) }^2 } {s_1},
\end{align*}
and by \citet{lattimore2020bandit}, we have
\begin{align*}
    \sum_{t=1}^T \min\set{ \nrm{ \phi\ind{t} }_{V_t^{-1}}^2, 1}
    \leq 2|\cI|\log\paren{1+\frac{TH}{|\cI|B}}.
\end{align*}
Combining the inequalities above and rescale $B\leftarrow \frac{B}{4|\cI|}$ completes the proof.
\end{proofof}

\section{Proofs from \cref{sec: PbRL}}\label{app: PbRL}
We present the full description of our algorithm or preference-based RL as follows.

\begin{algorithm}[t]
\caption{Outcome-Based Exploration for Preference-based RL}\label{alg: model-free-pbrl}
{\bfseries input:} Function class $\Fcal$, parameter $\lambda>0$, reference policy $\piexp$. \\
{\bfseries initialize:} $\Dcal \leftarrow \emptyset$. %
\begin{algorithmic}[1]
\For{$t = 1,2,\dotsc,T$}
\State Compute the optimistic estimates through \eqref{eq:max-f-R-pb}:
\begin{align*}
    (f^\iter{t},R^\iter{t})=\max_{f \in \cF, R \in \cR} \lambda \brac{ f_1(s_1) - \Vhref{\cD}{R}} - \LBE(f;R) - \LPBRM(R),
\end{align*}
\State Select policy $\pi^\iter{t} \leftarrow \pi_{f^\iter{t}}$. 
    \For{$h=1,2,\cdots,H$}
    \State Execute $\pi^\iter{t}\circ_h \piexp$ for two episode and obtain two trajectories $(\taua{t,h},\taub{t,h})$ and preference feedback $y^\iter{t,h}$.
    \State 
    \State Update dataset: $\Dcal \leftarrow \Dcal \cup \set{(\taua{t,h},\taub{t,h},y^\iter{t,h})}$.
    \EndFor
    
\EndFor
\State Output $\pihat = \unif(\pi^\iter{1:T})$.
\end{algorithmic}
\end{algorithm}

\subsection{Proof of \cref{thm: model-free-thm-pb}}\label{sec: proof-preference}

For each $t\in[T]$, we write $\cD\ind{t}$ to be the dataset maintained by \cref{alg: model-free} at the end of the $t$th iteration, i.e.,
\begin{align*}
\cD\ind{t}=\set{ (\taua{k,h},\taub{k,h},y^\iter{k,h}) }_{k\leq t, h\in[H]}.    
\end{align*}
Note that for each $t\in[T]$, $h\in[H]$, we have $\pib{t,h}=\piexp$. Therefore, for each $R\in\cR$, we define $\Vref{R}\defeq \EE^{\piexp}\brac{R(\tau)}$ and recall that
\begin{align*}
    \Vhref{\cD}{R}\defeq \frac{1}{|\cD|}\sum_{(\tau^+,\tau^-,y)\in\cD} R(\tau^-).
\end{align*}
The following lemma follows from the standard uniform convergence rate with Hoeffding's inequality and the union bound.
\begin{lemma}\label{lem:Vhat-concen-pb}
Let $\delta\in(0,1), \rho\geq 0$. Suppose that \cref{asmp:realizable} and \cref{asmp:complete} holds. 
Then \whp, for all $t\in[T], R\in\cR$, it holds that
\begin{align*}
    \abs{\Vhref{\cD^\iter{t}}{R}-\Vref{R}}\leq \sqrt{\frac{\log (2TN(\rho)/\delta)}{t}}+H\rho.
\end{align*}
\end{lemma}

We summarize the uniform concentration results for the loss $\LBE[\cD\ind{t}]$ and $\LPBRM[\cD\ind{t}]$ as follows. The proof is analogous to \cref{prop:all-loss-concen} and is provided in \cref{appdx:proof-all-loss-concen-pbrl}.
\begin{proposition}\label{prop:all-loss-concen-pbrl}
Let $\delta\in(0,1), \rho\geq 0$. Suppose that \cref{asmp:realizable} and \cref{asmp:complete} holds. 
Then \whp, for all $t\in[T], f\in\cF, R\in\cR$, it holds that
\begin{align*}
    &~ \abs{\Vhref{\cD^\iter{t}}{R}-\Vref{R}}\leq \sqrt{\frac{\kappa}{tH}}, \\
    &~\sum_{k\leq t} \sum_{h=1}^H \EE^{\pia{k,h},\pib{k,h}} \paren{ \brac{R(\tau^+)-R(\tau^-)} - \brac{\Rs(\tau^+)-\Rs(\tau^-)} }^2
    \leq C_\beta\brac{ \LPBRM[\cD^\iter{t}](R)-\LPBRM[\cD^\iter{t}](\Rsp) } +C_\beta H\kappa, \\
    &~\sum_{k\leq t} \sum_{h=1}^{H} \EE^{\pi^\iter{k}} \paren{ f_h(s_h,a_h)-[\BERh f_{h+1}](s_h,a_h)}^2
    \leq 2\brac{ \LBE[\cD^\iter{t}](f;R)-\LBE[\cD^\iter{t}](\Qsp;\Rsp) } +H\kappa,
\end{align*}
where $C_\beta=\frac{4e^{2\beta}}{\beta^2}$, 
\begin{align*}
    \kappa=C\paren{\log N(\rho)+\log(TH/\delta)+TH^2(\beta+1)(\epsapp^2+\rho)},
\end{align*}
and $C>0$ is an absolute constant.
\end{proposition}

\newcommand{\tR}{\widetilde{R}}
\newcommand{\tRs}{\tR^\star}
In the following, we condition on the success event of \cref{prop:all-loss-concen-pbrl}. Note that $\pib{t,h}\equiv\piexp$, and hence \cref{prop:all-loss-concen-pbrl} implies that for all $R\in\cR$, $t\in[T]$,
\begin{align*}
    \sum_{k\leq t} \sum_{h=1}^H \EE^{\pi^\iter{k}\circ_h\piexp} \paren{ \brac{R(\tau)-\Rs(\tau)} - \brac{\Vref{R}-\Vref{\Rs}} }^2
    \leq C_\beta\brac{ \LPBRM[\cD^\iter{t}](R)-\LPBRM[\cD^\iter{t}](\Rsp) } +C_\beta H\kappa.
\end{align*}
Therefore, for any reward function $R$, we define $\tR$ as $\tR_1(s,a)=R_1(s,a)-\Vref{R}$ and $\tR_h(s,a)=R_h(s,a)$ for $h>1$. Then it is clear that $\tR(\tau)=R(\tau)-\Vref{R}$, and for all $R\in\cR$, $t\in[T]$, we have
\begin{align}
    \sum_{k\leq t} \sum_{h=1}^H \EE^{\pi^\iter{k}\circ_h\piexp} \paren{ \tR(\tau)-\tRs(\tau)}^2
    \leq C_\beta\brac{ \LPBRM[\cD^\iter{t}](R)-\LPBRM[\cD^\iter{t}](\Rsp) } +C_\beta H\kappa.
\end{align}

\paragraph{Performance difference decomposition} In this setting, we re-write \eqref{eq:perf-diff-traj} as follows:
\begin{align*}
    f_1^\iter{t}(s_1)-V^{\pi^\iter{t}}(s_1)
    =&~ \sum_{h=1}^H \EE^{ \pi^\iter{t} }\brac{ f_h^\iter{t}(s_h,a_h)-[\BER[R^\iter{t}]f_{h+1}^\iter{t}](s_h,a_h) } \\
    &~+ \EE^{ \pi^\iter{t} }\brac{ \sum_{h=1}^H R^\iter{t}_h(s_h,a_h)-\sum_{h=1}^H\Rs_h(s_h,a_h) } \\
    =&~ \sum_{h=1}^H \EE^{ \pi^\iter{t} } e_h^\iter{t}(s_h,a_h) 
    + \EE^{ \pi^\iter{t} }\brac{ \tR^\iter{t}(\tau)-\tRs(\tau) } + \Vref{R^\iter{t}}-\Vref{\Rs},
\end{align*}
where we recall that we denote $e_h^\iter{t}\defeq f_h^\iter{t}-\BER[R^\iter{t}]f_{h+1}^\iter{t}$. Therefore, we re-organize the equality as
\begin{align}\label{eq:perf-diff-traj-pbrl}
\begin{aligned}
    \brac{f_1^\iter{t}(s_1)-\Vref{R^\iter{t}}}-\brac{V^{\pi^\iter{t}}(s_1)-\Vref{\Rs}}
    =&~ \EE^{ \pi^\iter{t} }\brac{ \tR^\iter{t}(\tau)-\tRs(\tau) }+\sum_{h=1}^H \EE^{ \pi^\iter{t} } e_h^\iter{t}(s_h,a_h).
\end{aligned}
\end{align}

With the above preparation, we present the proof of \cref{thm: model-free-thm-pb}, which closely follows the proof of \cref{thm: model-free-thm} in \cref{appdx:proof-cov-upper}.

\begin{proofof}{\cref{thm: model-free-thm-pb}}
By definition, for each $t\in[T]$,
\begin{align*}
    (f^\iter{t},R^\iter{t})=\max_{f \in \cF, R \in \cR} \lambda \brac{ f_1(s_1) - \Vhref{\cD^\iter{t-1}}{R} } - \LBE[\cD^\iter{t-1}](f;R) - \LPBRM[\cD^\iter{t-1}](R).
\end{align*}
Therefore, using $\Qsp\in\cF, \Rsp\in\cR$, we have
\begin{align*}
    &~\brac{ f_1^\iter{t}(s_1) - \Vhref{\cD^\iter{t-1}}{R^\iter{t}} } - \brac{ \Vsp_1(s_1) - \Vhref{\cD^\iter{t-1}}{\Rsp} } \\
    \leq&~ -\frac{\LBE[\cD^\iter{t-1}](f^\iter{t};R^\iter{t})-\LBE[\cD^\iter{t-1}](\Qsp;\Rsp)}{\lambda} - \frac{\LPBRM[\cD^\iter{t-1}](R^\iter{t})-\LPBRM[\cD^\iter{t-1}](\Rsp)}{\lambda}.
\end{align*}
Using the decomposition \eqref{eq:perf-diff-traj-pbrl}, \cref{prop:all-loss-concen-pbrl}, and the fact that $\abs{\Vsp_1(s_1)-\Vs_1(s_1)}\leq \epsapp$, $\abs{\Rsp(\tau)-\Rs(\tau)}\leq H\epsapp$, we have
\begin{align}
\begin{aligned}
    \Vs_1(s_1)-V^{\pi^\iter{t}}(s_1)
    \leq&~ (H+1)\epsapp+\abs{\Vref{R^\iter{t}}-\Vhref{\cD^\iter{t-1}}{R^\iter{t}}}+\abs{\Vref{\Rsp}-\Vhref{\cD^\iter{t-1}}{\Rsp}}+\frac{2H\kappa}{\lambda} \\
    &~+ \sum_{h=1}^H \paren{ \EE^{ \pi^\iter{t} }\brac{ e_h^\iter{t}(s_h,a_h)} - \frac{1}{C_\beta \lambda}\sum_{k<t} \EE^{\pi^\iter{k}} e_h^\iter{t}(s_h,a_h)^2 }\\
    &~+ \EE^{ \pi^\iter{t} }\brac{ \tR^\iter{t}(\tau)-\tRs(\tau) } - \frac{1}{2\lambda}\sum_{k<t} \sum_{h=1}^H \EE^{\pi^\iter{k}\circ_h \piexp} \paren{ \tR^\iter{t}(\tau) - \tRs(\tau) }^2.
\end{aligned}
\end{align}
Taking summation over $t=1,2,\cdots,T$ and apply \cref{prop:cov-Bellman-err}, \cref{prop:cov-RM-err}, and \cref{lem:Vhat-concen-pb} yields
\begin{align*}
    \sum_{t=1}^T \Vs_1(s_1)-V^{\pi^\iter{t}}(s_1)
    \leq \bigO{1}\cdot \brac{ H(\epsapp+\rho)+\sqrt{T\kappa} +  \frac{TH\kappa}{\lambda } + C_\beta \lambda H\ccov\log\paren{1+\frac{\ccov T}{\kappa}} }.
\end{align*}
This is the desired upper bound.
\end{proofof}

\subsection{Proof of \cref{prop:all-loss-concen-pbrl}}\label{appdx:proof-all-loss-concen-pbrl}

The inequality involving $\LBE[\cD^\iter{t}]$ is implied by \cref{prop:all-loss-concen} and proven in \cref{appdx:proof-all-loss-concen}. In the following, we only need to prove the inequality involving $\LPBRM[\cD^\iter{t}]$ by invoking \cref{prop:MLE-unif}.

Consider the class $\Theta=\cR\cup\set{\Rs}$, $\cX=(\cS\times\cA)^H\times (\cS\times\cA)^H$, and $\cY=\set{0,1}$. For any $R\in\Theta$, we define
\begin{align*}
    P_R(1|\tau^+,\tau^-)=\frac{\exp\paren{\beta R(\tau^+)}}{\exp\paren{\beta R(\tau^+)}+\exp\paren{\beta R(\tau^-)}}, \quad
    P_R(0|\tau^+,\tau^-)=\frac{\exp\paren{\beta R(\tau^-)}}{\exp\paren{\beta R(\tau^+)}+\exp\paren{\beta R(\tau^-)}},
\end{align*}
following \cref{def:BTR}. 

Recall that $\cD\ind{t-1}=\set{(\taua{k,h},\taub{k,h},y^\iter{k,h})}_{k<t,h\in[H]}$ 
is the history up to the $t$-th iteration. For simplicity, we denote $x^\iter{t,h}\defeq (\taua{t,h},\taub{t,h})$.
Note that $(x^\iter{t,1},y^\iter{t,1}),\cdots,(x^\iter{t,H},y^\iter{t,H})$.
Then it is clear that for all $t\in[T], h\in[H]$,
\begin{align*}
    \PP(y^\iter{t,h}|x^\iter{t,h}, \cD^\iter{t-1})=P_{\Rs}(y^\iter{t,h}|x^\iter{t,h}),
\end{align*}
and it also holds that
\begin{align*}
    L(R(\tau^+)-R(\tau^-),y)=-\log P_R(y|\tau^+,\tau^-), \qquad \forall y\in\set{0,1}.
\end{align*}
Further, noting that $\Nlog(\Theta,2H\beta\rho)\leq N(\cR,\rho)+1$.
Therefore, applying \cref{prop:MLE-unif} gives the following result: \whp[\frac{\delta}{2}], for any $R\in\cR$, $t\in[T]$,
\begin{align*}
    &~ \sum_{k=1}^t \sum_{h=1}^H \EE^{\pia{k,h},\pib{k,h}} \DHs{ P_R(\cdot|\tau^+,\tau^-), P_{\Rs}(\cdot|\tau^+,\tau^-)} \\
    \leq &~ 
    \frac12\sum_{(\tau^+,\tau^-,y)}\brac{ L(R(\tau^+)-R(\tau^-),y)-L(\Rs(\tau^+)-\Rs(\tau^-),y) } \\
    &~ + \log(N(\cR,H\rho)+1)+\log(2/\delta)+TH^3\beta\rho \\
    \leq&~ \frac12\brac{ \LPBRM[\cD^\iter{t}](R)-\LPBRM[\cD^\iter{t}](\Rsp) }+ \frac12 H\kappa,
\end{align*}
where the second inequality uses the fact that $\abs{\Rsp(\tau)-\Rs(\tau)}\leq H\epsapp$.
Finally, note that $\DHs{\Bern{p},\Bern{q}}\geq \frac{1}{2}(p-q)^2$ and 
\begin{align*}
    \abs{\frac{1}{e^{\beta w}+1}-\frac{1}{e^{\beta w'}+1}}\geq \frac{\beta}{2e^\beta}\abs{w-w'}, \qquad \forall w,w'\in[-1,1].
\end{align*}
Therefore, using the definition of $P_R$ completes the proof.
\qed

\section{Proofs of Lower Bounds}
\subsection{Hard Case of Learning with Fitted Reward Models}\label{sec: hard-case-separate}

As mentioned in \pref{sec: upper-bound-cov}, in \pref{alg: model-free} the learner has to optimize over the reward class and value function class jointly. In the following, we argue that if the learner first learns a fitted reward model in the reward class, then optimizes the value function with the fitted rewards, the output policies at each iteration never converge to the optimal policy. 

In detail, we consider algorithms in the form of \pref{alg:fitted-R}, where the learner fits the reward model $R\ind{t}$ at iteration $t$ first, then the learner calls algorithm $\alg$, which takes per-step rewards data as input and outputs a policy $\pi_t$ at each iteration. To align with the structure of \pref{alg: model-free}, we take $\alg$ to be a single iteration of the GOLF algorithm in \citet{jin2021bellman}, i.e. $\pi\ind{t} = \pi_{f\ind{t}}$ where $f\ind{t} = \argmax_{f\in \calF\ind{t}} f(x_1, \pi_f(x_1))$. Here the confidence set $\calF\ind{t}$ is defined as 
$$\calF\ind{t} = \left\{f\in \calF: \LBE[\cD\ind{t-1}](f;\widehat{R}\ind{t-1})\le \beta\right\}$$
with $\LBE$ defined in \pref{eq:loss-Bellman-full}.

\begin{algorithm}
\caption{RL with fitted reward models}\label{alg:fitted-R}
{\bfseries input:} Algorithm $\alg$, reward regression oracle $O$. 
\begin{algorithmic}[1]
\State \textbf{Initialize } $\calD_h\ind{0} = \emptyset$ for every $h\in [H]$
\For{$t = 1,2,\dotsc,T$}
\State  Feed $\calD\ind{t-1}$ to $\alg$ and receive $\pi^\iter{t}$ from $\alg$
\State Execute $\pi^\iter{t}$ and receive $(\tau^\iter{t},r^\iter{t})$, where $\tau\ind{t} = (s_1\ind{t}, a_1\ind{t}, \cdots, s_H\ind{t}, a_H\ind{t})$
\State Receive the fitted reward function from $O$:
\begin{align*}
\widehat{R}^\iter{t}=\min_{R \in \cR} \sum_{k=1}^t (R(\tau^\iter{k})-r^\iter{k})^2.
\end{align*}
\State Let $\widehat{r}_h\ind{t}=\widehat{R}\ind{t}_h(s_h\ind{t},a_h\ind{t})$ for each $h\in[H]$.
\State Let $\calD\ind{t} = \calD\ind{t-1}\cup \{(s_1\ind{t}, a_1\ind{t}, \widehat{r}_1\ind{t},\cdots,s_H\ind{t}, a_H\ind{t}, \widehat{r}_H\ind{t})\}$.
\EndFor
\end{algorithmic}
\end{algorithm}

Then we have the following proposition, which shows that this approach outputs suboptimal policies at every iteration in some special hard cases.

\begin{proposition}\label{prop: golf-lower-bound}
    Consider \pref{alg:fitted-R} with $\alg$ to a single iteration of the GOLF algorithm. After running $T$ iterations, the learner averages over all policies to output a policy. There exists an MDP class that realizes the ground truth MDP, such that the above algorithm outputs a policy which is at least $0.01$-suboptimal.
\end{proposition}

\cfcomment{the notation below is a bit confusing}
\begin{proof}[\cpfname{prop: golf-lower-bound}]
    We consider the following class of two-layer MDP, where $\calS_1 = \{s_1\}$, $\calS_2 = \{s_2\}$, and the action space to be $\calA = \{a_1, a_2\}$. The transition models $\TT$ are identical across the class, and have the following form:
    $$\TT(s_2\mid s_1, a_i) = 1,\qquad \forall i \in \{1, 2\}.$$
    The reward class is defined as $\calR = \{R^1, R^2\}$, where 
    \begin{align*} 
        & R^1(s_1, a_1) = R^1(s_1, a_2) = 0.20,\quad R^1(s_2, a_1) = 0.20,\quad R^1(s_2, a_2) = 0.19,\\
        \text{and}\quad & R^2(s_1, a_1) = R^2(s_1, a_2) = 0.00,\quad R^2(s_2, a_1) = 0.38,\quad R^2(s_2, a_2) = 0.39.
    \end{align*}
    The $Q$-function class $\calQ$ is defined as $\calQ = \{Q^1, Q^2, Q^3, Q^4
    \}$, which takes value in \pref{tab: q-value} respectively.
    \begin{table}[!htp]
        \caption{Value of $Q^1, Q^2, Q^3 , Q^4$}\label{tab: q-value}
        \centering
        \begin{tabular}{|c|c|c|c|c|}
            \hline
            & $(s_1, a_1)$ & $(s_1, a_2)$ & $(s_2, a_1)$ & $(s_2, a_2)$\\
            \hline
            $Q^1$ & $0.40$ & $0.40$ & $0.20$ & $0.19$ \\
            \hline
            $Q^2$ & $0.20$ & $0.20$ & $0.20$ & $0.19$\\
            \hline
            $Q^3$ & $0.59$ & $0.59$ & $0.38$ & $0.39$ \\
            \hline
            $Q^4$ & $0.39$ & $0.39$ & $0.38$ & $0.39$\\
            \hline
        \end{tabular}
    \end{table}
    Notice that in all possible reward models and $Q$-functions, the values at $(s_1, a_1)$ and at $(s_1, a_2)$ are the same. In the following, when without ambiguity we simply use $R(s_1)$ to denote $R(s_1, a_1)$ and $R(s_1, a_2)$, and use $Q(s_1)$ to denote $Q(s_1, a_1)$ and $Q(s_2, a_2)$.
    
We further suppose the ground truth model reward satisfies $R = R^1$, then we can verify that the optimal $Q$-function is $Q^1$. It is easy to verify that sets $\calQ$ and $\calR$ satisfy the completeness assumption. Hence, sets $\calQ$ and $\calR$ satisfy the realizability assumption \pref{asmp:realizable} and the completeness assumption \pref{asmp:complete} with $\calG = \calQ$. 

    To see why this is a hard-case for GOLF type algorithms, we first notice that for any trajectory $\tau = (s_1, \ta_1, s_2, \ta_2)$ with outcome reward $r = R^1(s_1, \ta_1) + R^1(s_2, \ta_2)$ collected by the algorithm, we always have 
    $$r = R^2(s_1) + R^2(s_2, \ta_2).$$
    Hence as long as $\calD$ does not contain state-action pair $(s_2, a_1)$, when fitting the reward function using the following ERM oracle:
    $$R = \argmin_{R\in \calR} \sum_{(\tau, r)\in \calD} (r(\tau) - r)^2,$$
    the reward model $R^2$ always achieves the minimum. In the worst case, we assume the fitted reward models encountered by the learner at such rounds are always $R^2$.

    In the following, we verify that by running the GOLF algorithm, the learner will not encounter the state-action pair $(s_2, a_1)$ at any round. We notice that the optimal policies of $Q^3$ and $Q^4$ all take $a_2$ at state $s_2$, and also that
    $$Q^3(s_1)\ge Q^1(s_1)\quad \text{and}\quad Q^3(s_1)\ge Q^2(s_1).$$
    Hence, to verify that the algorithm never chooses $a_1$ at state $s_2$, we only need to verify that if either $Q^1$ or $Q^2$ belongs to the confidence set, then $Q^3$ also belongs to the confidence set. 

    When the learner collects a new trajectory, two new pieces of data will be added to the dataset $\calD$. If the trajectory does not pass through the state-action pair $(s_2, a_1)$, these two pieces of data will be in the following form:
    $$(s_1, a_1, R^2(s_1)), \quad (s_2, a_2, R^2(s_2, a_2))\quad \text{or}\quad (s_1, a_2, R^2(s_1)),\quad (s_2, a_2, R^2(s_2, a_2)).$$
    No matter which one of these two, we have the following inequality for the sum of squared Bellman error across these two pieces of data
    \begin{align*}
        \calE_1(Q^1)^2 + \calE_2(Q^1)^2 & = 0.20^2 + 0.20^2\ge 0.20^2 = \calE_1(Q^3)^2 + \calE_2(Q^3)^2,\\
        \calE_1(Q^2)^2 + \calE_2(Q^2)^2 & = 0.01^2 + 0.28^2\ge 0.20^2 = \calE_1(Q^3)^2 + \calE_2(Q^3)^2.
    \end{align*}
    According to the construction of the confidence set, if either $Q^1$ or $Q^2$ belongs to the confidence set, then $Q^3$ belongs to the confidence set as well.

    Therefore, no matter how many rounds the algorithm runs, the optimistic policy always takes action $a_2$ at state $s_2$. Hence the average policy $\pihat$ also takes $a_2$ at $s_2$, which implies that
    $$J(\pi^\star) - J(\pihat)\ge 0.01.$$
    
\end{proof}

\subsection{Proof of \pref{thm: lower-bound}}\label{app: exp-lower}
\newcommand{\B}{\mathbf{B}}
\newcommand{\indic}[1]{\mathbf{1}\set{#1}}

Fix a parameter $\epsilon\in(0,1)$ and $N\leq \paren{\frac1{2\epsilon}}^{d/2}$. Then, by the standard packing argument over sphere \citep[see e.g.,][]{li2022understanding}, there exists a set $\Theta=\set{\theta_1,\cdots,\theta_N}\subseteq \mathbb{S}^{d-1}$ such that
\begin{align*}
    \nrm{\theta_i-\theta_j}\geq \sqrt{2\epsilon}, \qquad\forall i\neq j.
\end{align*}
This implies $\lr \theta_i,\theta_j\rr\leq 1-\epsilon$ for any $i\neq j$.

\paragraph{Construction}
In the following, we set $b=1-\epsilon$, and construct state space $\cS$ as
\begin{align*}
    \calS = \calS_1 \sqcup \calS_2, \qquad
    \calS_1 = \{s_1\}, \qquad
    \calS_2 = \Theta,
\end{align*}
and let action space $\cA=\Theta$. The initial state is always $s_1$, and we define the transition $\TT$ as
\begin{align*}
    \TT(s_2=\theta \mid s_1, a=\theta) = 1, \qquad\forall \theta\in\Theta,
\end{align*}
i.e., taking action $a=\theta$ at $s_1$ transits to $s_2=\theta$ deterministically. 

\paragraph{Reward functions}
For any $v\in\Theta$, we define the reward model $R^{v}$ as follows:
\begin{align*} 
    R^{v}_1(s, a) & = \frac13\brac{\relu(\langle a, v\rangle - b) + \langle a, v\rangle+1}\in[0,1], & \forall a\in\Theta,\\
    R^{v}_2(s,a) & =  \frac13\brac{1-\langle s, v\rangle}\in[0,1],  & \forall s\in\Theta.
\end{align*}
Note that we can write $g_1(x)=\frac{1}{3}\brac{\relu(x-b)+x+1}, g_2(x)=\frac{1-x}{3}$, and then $g_2$ is a linear function, and
\begin{align*}
    \frac13\abs{x-y}\leq \abs{g_1(x)-g_2(y)}\leq \frac23\abs{x-y}, \qquad\forall x,y\in\R.
\end{align*}
Hence, $g_1$ and $g_2$ are (well-conditioned) generalized linear functions, and hence $R_1^v$ and $R_2^v$ are both (well-conditioned) $d$-dimensional generalized linear functions.

We let $M^v$ be the MDP with transition $\TT$ and mean reward function $R^v$, and $\cM=\set{M^v:v\in\Theta}$ be the corresponding class of MDPs.
We next show that $\calM$ can be learned with polynomial process-based samples, but cannot be learned with polynomial outcome-based samples.

\paragraph{Exponential Lower Bound for Outcome-Based Setting} 
When executing a policy $\pi$ in MDP $M^{v}$, we have $a_1=\pi_1(s_1)$, $s_2=a_1$, and $a_2=\pi_2(s_2)$, and the data $(\tau_{\theta,\theta'},R)$ observed are in the following form of trajectory together outcome-based rewards:
\begin{align*}
    \tau_{\pi}=(s_1, a_1, s_2, a_2), \qquad R|\tau_{\pi}\sim \Bern{ \frac13\relu(\langle a_1, v\rangle - b)+\frac23 },
\end{align*}
where $\tau_{\pi}$ is a deterministic function of $\pi$. In the following, we denote $a_\pi=\pi_1(s_1)$, and then $\EE[R|\pi]=\frac14\relu(\langle a_\pi, v\rangle - b)+\frac12$. Further, under $M^v$,
\begin{align*}
    J(\pi)=\frac{2}{3}+\frac\epsilon3\indic{a_\pi=v},
\end{align*}
and in particular, $J(\pis)=\frac23+\frac{\epsilon}{3}$. Therefore, for any policy $\pi$, it is $(\epsilon/3)$-optimal under $M^v$ only when $a_\pi=v$. Hence, we can apply the standard lower bound argument for multi-arm bandits~\citep[see e.g.,][]{lattimore2020bandit} to show that: If there any $T$-round algorithm that returns an $(\epsilon/3)$-optimal policy with probability at least $\frac34$ for any MDP $M^v\in\cM$, then it must hold that $T\geq c\frac{N}{\epsilon^2}$ (where $c>0$ is an absolute constant). Setting $\epsilon=\frac{1}{3}$ completes the proof of the lower bound.
\qed

\paragraph{Polynomial Upper Bound with Process-Based Samples} 
Notice that for fixed $v\in\Theta$, under $M^v\in\cM$, we have
\begin{align*}
    J(\pis)-J(\pi)=\frac\epsilon3\brac{1-\indic{a_\pi=v}}=\frac13\brac{\lr v,v\rr - \lr a_\pi,v\rr}.
\end{align*}
Thus, for any $\theta\in\Theta$, we define $\pi^\theta$ as $\pi^\theta_1(s)=\pi^\theta_2(s)=\theta$ for $\forall s\in\cS$. Then it holds that under $M^v$,
\begin{align*}
    \EE\cond{\frac{1}{3}-R_2}{\pi^\theta}=\frac13\langle \theta, v\rangle.
\end{align*}
Therefore, given access to process reward feedback, we can reduce learning $\cM$ to learning a class of linear bandits. Hence, for any $\alpha>0$, with process reward feedback, there are algorithms that returns an $\alpha$-optimal policy with high probability, using $T\leq \tbO{d^2/\alpha^2}$ episodes with process rewards~\citep[see e.g.,][]{dani2008stochastic}.
\qed

\end{document}